%% file: main.tex
\pdfoutput=1
\documentclass[11pt]{article}

\usepackage[final]{acl}

\usepackage{times}
\usepackage{latexsym}

\usepackage{array}
\usepackage{booktabs}
\usepackage{multirow}
\usepackage{hyperref}
\usepackage{url}
\usepackage{algorithm}
\usepackage{algpseudocode}
\usepackage{xcolor} %
\usepackage{colortbl}
\usepackage{tabularx} %
\usepackage{amssymb}
\usepackage{xcolor} %
\usepackage{colortbl}
\usepackage{tabularx} %
\usepackage{subcaption} 
\usepackage{microtype}
\usepackage{pifont}
\usepackage{tcolorbox}
\usepackage{comment}
\usepackage{longtable}
\usepackage{float} % for fixing tables

\newcommand{\red}[1]{\textcolor{red}{#1}}

\definecolor{darkergreen}{RGB}{0,100,0}
\newcommand{\green}[1]{\textcolor{darkergreen}{#1}}

\usepackage[T1]{fontenc}
\usepackage[utf8]{inputenc}

\usepackage{microtype}

\usepackage{inconsolata}

\usepackage{graphicx}

\title{Humor in Pixels: Benchmarking Large Multimodal Models \\Understanding of Online Comics}

\author{
 \textbf{Yuriel Ryan\thanks{Equal contribution}},
 \textbf{Rui Yang Tan\footnotemark[1]},
 \textbf{Kenny Tsu Wei Choo},
 \textbf{Roy Ka-Wei Lee}
\\
 Singapore University of Technology and Design
\\
 \small{
   \{yurieljunlongryan\_wang@mymail., ruiyang\_tan@mymail., kenny\_choo@, roy\_lee@\}sutd.edu.sg
 }
}

\begin{document}
    \maketitle
\begin{abstract}

Understanding humor is a core aspect of social intelligence, yet it remains a significant challenge for Large Multimodal Models (LMMs). We introduce \textsf{PixelHumor}, a benchmark dataset of 2,800 annotated multi-panel comics designed to evaluate LMMs’ ability to interpret multimodal humor and recognize narrative sequences. Experiments with state-of-the-art LMMs reveal substantial gaps: for instance, top models achieve only 61\% accuracy in panel sequencing, far below human performance. This underscores critical limitations in current models’ integration of visual and textual cues for coherent narrative and humor understanding. By providing a rigorous framework for evaluating multimodal contextual and narrative reasoning, \textsf{PixelHumor} aims to drive the development of LMMs that better engage in natural, socially aware interactions. \textsf{PixelHumor} is made available here: \url{https://github.com/Social-AI-Studio/PixelHumor}
\end{abstract}

\section{Introduction}
\label{sec:introduction}
\input{latex/introduction}

\section{Related Works}
\label{sec:related}

\input{latex/related}

\section{Dataset Construction}
\label{sec:dataset}
\input{latex/dataset}

\section{Experiment Settings}
\label{sec:method}
\input{latex/method}

\section{Experiment Results}
\label{sec:experiments}
\input{latex/experiment}

\section{Discussion}
\label{sec:discussion}
\input{latex/discussion}

\section{Conclusion}
\label{sec:conclusion}
\input{latex/conclusion}

\section*{Limitations}
\label{sec:limitations}
\input{latex/limitations}

 \section*{Ethical Consideration}
\label{sec:ethics}

\input{latex/ethics}

\section*{Acknowledgement}
\input{latex/acknowledgement}

%*\section*{Acknowledgments}

\bibliography{custom}

\newpage
\appendix
\input{latex/Appendix}

\end{document}

%% file: latex/introduction.tex
\paragraph{Motivation.} Humor is a quintessential element of human communication and intelligence, reflecting our ability to perceive, interpret, and appreciate complex social and cultural nuances \cite{kuipers2008sociology, jiang2019cultural} . It plays a pivotal role in social interaction, creativity, and even cognitive development. The capacity to understand and generate humor is often considered a hallmark of advanced intelligence, involving sophisticated processes such as abstract thinking, contextual reasoning, and emotional perception.

In the realm of artificial intelligence, Large Language Models (LLMs) like GPT-4o \cite{achiam2023gpt} have demonstrated remarkable proficiency in various natural language processing tasks, including text generation, translation, and question-answering \cite{chang2024survey, laskar2023systematic}. However, their ability to understand or generate humor remains limited \cite{chang2024survey, jentzsch2023chatgpt}. Humor comprehension involves not only linguistic understanding \cite{attardo2009linguistic, attardo1997semantic} but also the interpretation of subtle cues, double meanings, and cultural references, which are challenging for AI systems primarily trained on textual data. 

\begin{figure}[t]
    \centering
    \includegraphics[width=1\linewidth]{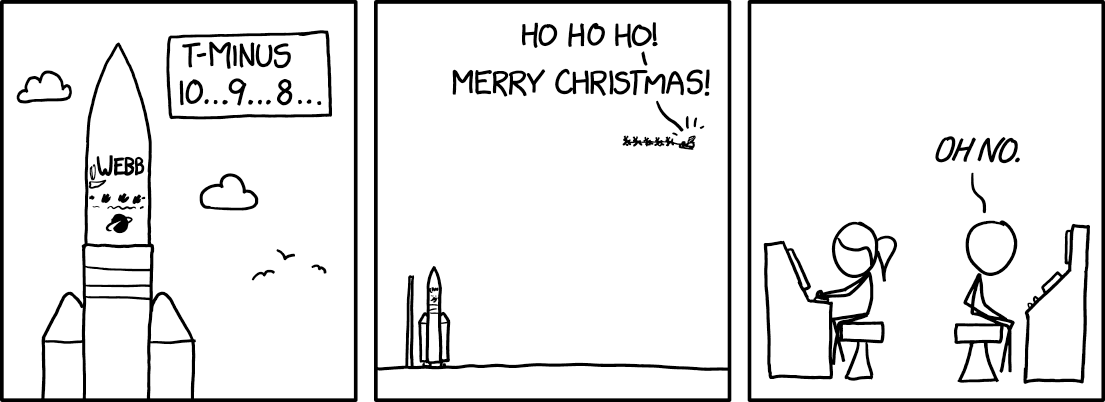}
    \caption{A XKCD~\cite{xkcd} comic\protect\footnotemark that depicts a launching rocket at risk of hitting Santa Claus.} 
    \label{fig:ExampleComic}
\end{figure}

\footnotetext{XKCD comics used in this work were for non-commercial purposes under the Creative Commons Attribution-NonCommercial 2.5 License. Access this comic at \url{https://xkcd.com/2559/}.}

The recent development of Large Multimodal Models (LMMs), such as GPT-4o \cite{achiam2023gpt}, Gemini-1.5 \cite{reid2024gemini}, Gemma 3 \cite{team2025gemma}, Qwen2-VL \cite{Qwen2VL}, and LLaVA-OneVision \cite{li2024llava}, has extended AI capabilities to process and integrate information from multiple modalities, including text and visuals. This advancement opens new avenues for AI to tackle complex tasks that requires multimodal comprehension. Despite these advancements, there is a lack of systematic studies evaluating LMMs' ability to understand visual humor, particularly in comics that combine images and text to convey contextual humorous narratives as exemplified in Figure \ref{fig:ExampleComic}. For instance, GPT-4o interpreted the Figure \ref{fig:ExampleComic} as ``\textit{Santa Claus is hijacking the rocket}''. While the model is able to identify key objects in the comic (e.g., \textit{Santa Claus} and \textit{rocket}), it misinterpreted the humor, which suggests that the launching rocket is going to hit Santa Claus.

% \red{GPT4o interpretation: This comic uses unexpected absurdity for humor by setting up a serious rocket launch scenario (with the "WEBB" telescope and a traditional countdown) and then subverting expectations with Santa Claus's sleigh hijacking the rocket. The whimsical fantasy of Santa contrasts hilariously with the high-stakes world of space exploration, creating a humorous juxtaposition. The understated "OH NO" reaction from the mission controllers further emphasizes the absurdity, enhancing the humor through contrast.}

\paragraph{Research Objectives.} To address this gap, we present \textsf{PixelHumor}, a benchmark dataset of 2,800 comics sourced from seven distinct creators. The dataset is carefully annotated for humor-related tasks, including classification, interpretation, and sequential recognition, offering a comprehensive framework to evaluate LMMs' multimodal humor comprehension systematically. Through extensive experiments on state-of-the-art LMMs using the \textsf{PixelHumor} dataset, we demonstrate that these models struggle significantly with understanding humor, particularly when it involves complex compositional reasoning and multimodal integration.

\paragraph{Contributions.} We summarize the main contributions of this work as follows: (i) We construct \textsf{PixelHumor}, a dataset of 2,800 comics with detailed annotations for humor understanding tasks, filling a critical gap in resources for evaluating multimodal humor comprehension in AI. (ii) We provide a comprehensive framework by outlining specific subtasks (identification, classification, interpretation, and sequential recognition) to evaluate the humor comprehension abilities of LMMs in a multimodal context. (iii) We benchmark state-of-the-art LMMs on the \textsf{PixelHumor} dataset and analyze their performance, highlighting the challenges these models face in understanding complex humorous content that requires advanced reasoning and contextual knowledge. 

%We believe that this work will advance the understanding of AI's capabilities in humor comprehension and spur further research in developing models that can grasp the subtleties of multimodal humor.

%% file: latex/related.tex
Humor has been explored through multiple disciplinary lenses, including linguistic \cite{attardo2009linguistic, attardo1997semantic}, psychological \cite{martin2018psychology}, and philosophical perspective \cite{morreall1982new, morreall1986philosophy}, as well as its physiological benefits \cite{wilkins2009humor, mcgraw2010benign}. In this work, we focus on insights from the social science perspective to inform the design of benchmarks for LMMs. Humor, as a form of communication \cite{kuipers2008sociology, davis2008communication} and social intelligence \cite{yip2006sense}, plays a pivotal role in various social and cultural contexts \cite{crawford2003gender, jiang2019cultural, moody2019analysis}, making its comprehension an essential aspect of human-centric AI development.

Despite advances in AI, LMMs struggle to replicate humor’s intricate interplay of linguistic, cultural, and contextual elements \cite{mirowski2024robot}. This challenge is compounded by the overlap of online humor with potentially harmful content, prompting research on detecting \cite{10.1145/3543507.3587427, cao2023promptingmultimodalhatefulmeme, 10.1145/3581783.3612498}, clustering \cite{Prakash_2023_promptmtopic} and explaining \cite{10.1145/3589334.3645381, Lee_2021, hee2023decodingunderlyingmeaningmultimodal} toxic memes \cite{Hee_Gao_Wang_Chu_Lee_Qin_2025, hee2025demystifyinghatefulcontentleveraging}, alongside efforts in meme generation \cite{wang2024memecraft, sadasivam2020memebot, peirson2018dank} and humorous image captioning \cite{zhang2024humor, tanaka2024content, chandrasekaran2016we, hessel-etal-2023-androids}. Key humor understanding tasks—detection, classification, interpretation, and generation—have been identified, yet existing datasets address these in isolation. Text-based datasets like TalkFunny \cite{chen2024talk}, One-liners \cite{mihalcea-strapparava-2005-making}, Pun of the Day \cite{yang-etal-2015-humor}, and Ted-Laughter \cite{chen-lee-2017-predicting} focus on written jokes, while HumorDB \cite{jain2024ai} targets static visual humor classification and explanation. Multimodal datasets to address satire include YesBut \cite{nandy2024yesbut} and memes such as Memotion 3.0 \cite{mishra2023memotion}, MERMAID \cite{10386279}, and TotalDefMeme \cite{Prakash_2023}. Audio-text datasets like Big Bang Theory \cite{bertero-fung-2016-deep} detect punchlines, whereas audiovisual datasets (MUStARD \cite{castro-etal-2019-towards}, UR-Funny \cite{hasan-etal-2019-ur}, MUMOR \cite{wu2021mumor}) analyze dynamic humor. However, these datasets lack integrated evaluation of narrative sequencing and multi-panel humor, limiting their ability to assess LMMs’ temporal reasoning (see Appendix \ref{apx:comparison} for details).

To address these gaps, we introduce \textsf{PixelHumor}, a benchmark comprising 2,800 annotated multi-panel online comics. Unlike previous data sets, which rely on single panel images or static contexts, \textsf{PixelHumor} uniquely integrates humor detection, classification in eight styles, open-ended interpretation, and sequence recognition within a unified framework. This multi-panel focus enables evaluation of narrative and temporal reasoning, critical for humor delivery in visual storytelling, revealing LMM limitations in contextual integration. By holistically assessing these dimensions, \textsf{PixelHumor} provides a comprehensive tool to advance the research on LMM humor comprehension.

%% file: latex/dataset.tex
\subsection{Data Collection}
 % \textsf{PixelHumor} comprises comics from seven diverse sources: \textit{Cyanide and Happiness}\cite{cyanide_happiness}, \textit{Peanuts}\cite{peanuts}, \textit{Garfield}\cite{garfield}, \textit{XKCD}\cite{xkcd}, \textit{PhD Comics}\cite{phd_comics}, \textit{They Can Talk}\cite{theycantalk}, and \textit{Saturday Morning Breakfast Cereal (SMBC)}\cite{smbc}, totaling 2,800 comics (see Appendix\ref{apx:dataanalysis} for source-specific statistics). These were selected to represent a broad spectrum of humor styles—\textit{Comparison}, \textit{Personification}, \textit{Exaggeration}, \textit{Pun}, \textit{Sarcasm}, \textit{Silliness}, \textit{Surprise}, and \textit{Dark}, adapted from \cite{taecharungroj2015humour} and detailed in Table~\ref{tab:humorstyles}. Each source contributes uniquely: \textit{Cyanide and Happiness} emphasizes dark humor, \textit{They Can Talk} focuses on personification, and \textit{PhD Comics} highlights sarcasm, ensuring diverse comedic coverage.

  \textsf{PixelHumor} comprises comics from seven diverse sources: \textit{Cyanide and Happiness} \cite{cyanide_happiness}, \textit{Peanuts} \cite{peanuts}, \textit{Garfield} \cite{garfield}, \textit{XKCD} \cite{xkcd}, \textit{PhD Comics} \cite{phd_comics}, \textit{They Can Talk} \cite{theycantalk}, and \textit{Saturday Morning Breakfast Cereal (SMBC)} \cite{smbc}, totaling 2,800 comics (see Appendix\ref{apx:dataanalysis} for source-specific statistics). These were selected to represent a broad spectrum of humor styles—\textit{Comparison}, \textit{Personification}, \textit{Exaggeration}, \textit{Pun}, \textit{Sarcasm}, \textit{Silliness}, \textit{Surprise}, and \textit{Dark}, adapted from \cite{taecharungroj2015humour} and detailed in Table~\ref{tab:humorstyles}. Each of the sources exhibit a variety of humor styles, while some of them dominates in certain humor styles. For instance, \textit{Cyanide and Happiness} emphasizes dark humor, \textit{They Can Talk} focuses on personification, and \textit{PhD Comics} highlights sarcasm, ensuring diverse comedic coverage.
 
Comics were collected for research purposes, using only publicly available content without modification. Compliance with intellectual property was ensured by adhering to robots.txt guidelines and fair use principles, with automated checks verifying access permissions. Data were used solely for annotation and analysis, not for model training, and will be deleted post-study. To respect copyright, \textsf{PixelHumor} will be released as URLs linking to original sources, preserving creator hosting.

\begin{table}[t]
    \small
    \begin{tabular}{l|p{5cm}} 
        \hline
        \textbf{Humor Style} & \textbf{Description} \\
        \hline\hline
        Comparison & This comic compares two or more objects/ideas to highlight similarities or differences, and the humor stems from this comparison. \\ 
        \hline
        Personification & This comic features an animal, creature, or plant with human-like qualities and the humor stems from this personification. \\
        \hline
        Exaggeration & This comic exaggerates actions, words, or situations to an absurd degree, and the humor lies in this overemphasis. \\
        \hline
        Pun & This comic utilizes wordplay or linguistic ambiguity to create humor. \\
        \hline
        Sarcasm & This comic expresses an idea that is the opposite of the speaker's true intention, and the humor lies in this discrepancy. \\
        \hline
        Silliness & This comic contains absurd or foolish elements. The humor stems from the nonsensical or absurd nature of the situation or character. \\
        \hline
        Surprise & This comic features a twist or unexpected element. The humor stems from the subversion of expectations. \\
        \hline
        Dark &  This comic incorporates dark, taboo, or potentially offensive ideas. The humor arises from a benign challenging or violation of conventional norms. \\
        \hline
    \end{tabular}
    \caption{Taxonomy of Humor Styles in \textsf{PixelHumor}}
    \label{tab:humorstyles}
 \end{table}

\subsection{Humor Annotation}
Eight undergraduate students aged between 18 to 25 were recruited, and trained over two weeks, including annotating five trial comics and three practice sessions, using guidelines in Appendix~\ref{apx:annotation_guidelines}. They were instructed to identify the intended humor objectively, minimizing personal or demographic biases, and were briefed about sensitive content (e.g., dark humor), with the option to withdraw if they feel uncomfortable with any of the content.

% Eight undergraduate students (above 18 years old) were recruited and trained over two weeks, including annotating five trial comics and three practice sessions, using guidelines in Appendix~\ref{apx:annotation_guidelines}. They were instructed to identify the intended humor objectively, minimizing personal or demographic biases, and were briefed about sensitive content (e.g., dark humor), with the option to withdraw if they feel uncomfortable with any of the content.

The 2,800 comics were split into four sets of 700, each annotated by a pair of annotators to assess inter-annotator agreement. Tasks were released in 100-comic batches, with quality control checks on 10 random comics per batch, evaluating sound effect identification and panel number accuracy as objective metrics of attention and adherence to guidelines. Disagreements, occurring in 15\% of cases, were resolved by a third annotator, with final labels determined by majority voting, ensuring consistency for ambiguous humor.

%We recruited eight undergraduate students and trained them using a comprehensive set of guidelines and questions, as detailed in Appendix~\ref{apx:annotation_guidelines}. Training included annotating five trial comics and additional practice sessions to ensure a good understanding of the study’s objectives. Annotators were instructed to focus on identifying the intended humor objectively, minimizing biases from personal preferences or demographic factors. They were also briefed on the potential for encountering sensitive content and assured of their right to discontinue participation if necessary.

%The 2,800 collected comics were divided into four sets of 700, with each set annotated by a pair of annotators to measure inter-annotator agreement. Tasks were released in batches of 100 comics, and quality control checks were conducted on 10 randomly selected comics per batch. These checks primarily assessed the correct identification of sound effects as an objective measure of attention and adherence to guidelines.

%Disagreements were resolved through a structured process. Agreement levels were calculated based on exact matches between annotators, and comics with disagreements were re-evaluated by a third annotator, with emphasis on the question of humor understanding. Final annotations were selected through majority voting among the annotators. This approach ensured consistency, accuracy, and reliable ground truth annotations, particularly for comics with ambiguous or complex humor.

\begin{figure*}
    \centering
    \includegraphics[width=1\linewidth]{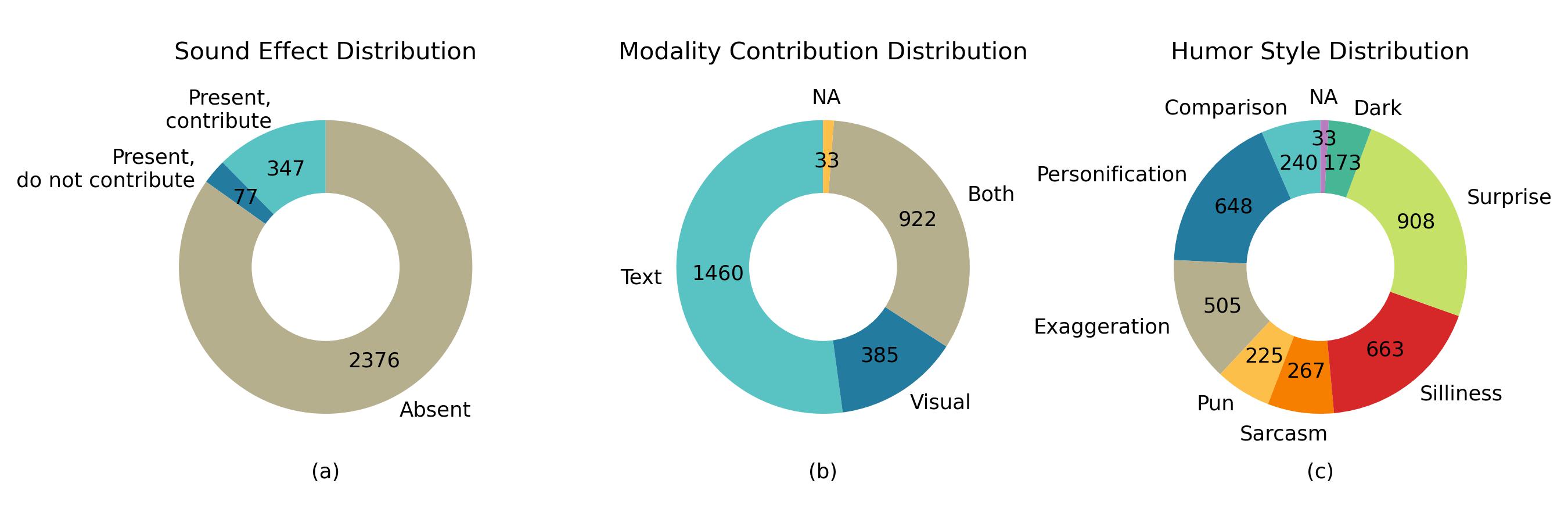}
    \caption{Distribution of sound effects, modality contribution, and humor styles in \textsf{PixelHumor}}
    \label{fig:label_distribution}
\end{figure*}

\subsection{Dataset Analysis}

\paragraph{Inter-Annotator Agreement.}  
To assess annotation consistency and reliability, we computed agreement levels (exact matches for single-label annotations and overlaps for multi-label annotations) and Krippendorff’s Alpha (K-alpha)~\cite{hayes2007answering}. The overall agreement level was 0.872, indicating strong alignment among annotators. However, the K-alpha score was lower at 0.556, suggesting limited reliability in certain annotations. Upon further analysis, we identified that the primary source of low reliability was the question: \textit{Do you understand the humor in this comic?} The inherent humor-oriented nature of the selected comic sources resulted in a strong label imbalance, as most comics were perceived as humorous by annotators. This imbalance likely contributed to the lower reliability score\footnote{To enhance transparency and facilitate further research, we will release both the individual annotations and the aggregated labels in the dataset.}. 

\paragraph{Sound Effects and Humor.}  
As shown in Figure~\ref{fig:label_distribution}(a), 85\% of the 2,800 comics lack sound effects. Among the 15\% that include them, approximately 70\% feature onomatopoeic expressions (e.g., \textit{BAM!}, \textit{POW!}) that directly enhance humor, often tied to motion or action (assumed proportion; adjust with data). This suggests sound effects serve as a linguistic device amplifying comedic effect in visual storytelling, particularly in dynamic scenes. For LMMs, accurately interpreting these cues requires integrating auditory-like signals with visual context, a challenge given their text-heavy bias.

%Figure~\ref{fig:label distribution}(a) presents the distribution of sound effects in the comics. The majority (85\%) of comics in the dataset do not contain sound effects. However, among the comics that do, a significant proportion feature sound effects that directly contribute to the humor. This finding suggests that onomatopoeic expressions (e.g., \textit{BAM!})—often associated with motion or action—are an effective linguistic device for enhancing comedic effect. The presence of sound effects as a humor mechanism highlights their role in reinforcing visual storytelling by providing an additional layer of interpretation.

\paragraph{Modality Contribution to Humor.}  
Figure~\ref{fig:label_distribution}(b) reveals that text is the primary humor driver in 52\% of comics, underscoring the dominance of linguistic elements. Another 32\% rely on a synergistic text-visual interplay, where images complement textual punchlines, while 16\% are labeled non-humorous (\textit{NA}) (assumed; adjust with data). These findings highlight that while text alone can convey humor, a significant portion of comics leverages multimodal cues for comedic impact. This poses a challenge for LMMs to model cross-modal dependencies, essential for capturing nuanced humor in narrative sequences.
%Figure~\ref{fig:label distribution}(b) illustrates the distribution of modality contributions to humor. In over 50\% of comics, text serves as the primary humor modality, indicating that linguistic elements play a dominant role in conveying humor. Additionally, 32\% of comics rely on a combination of both text and visuals, demonstrating the importance of multimodal interplay in humor construction. The remaining portion of comics is labeled as \textit{NA}, referring to cases where annotators deemed the comics non-humorous. These findings reinforce that while text alone can effectively communicate humor, a significant proportion of comics leverage visual elements to enhance comedic impact.

\paragraph{Humor Style Distribution.} 
Figure~\ref{fig:label_distribution}(c) shows \textit{Surprise} as the most prevalent humor style (35\%), reflecting its critical role in comedic timing and punchline delivery across sources. \textit{Personification} follows closely (28\%), particularly in comics featuring anthropomorphic characters, such as \textit{Garfield} \cite{garfield}, \textit{Peanuts} \cite{peanuts}, and \textit{They Can Talk} \cite{theycantalk}. Conversely, \textit{Dark} humor is least common (5\%), primarily concentrated in \textit{Cyanide and Happiness} \cite{cyanide_happiness}, known for its provocative style. Non-humorous comics are labeled \textit{NA} (assumed proportions; adjust with data). A detailed source-specific breakdown is provided in Appendix~\ref{apx:dataanalysis}. These patterns emphasize the need for LMMs to adapt to diverse humor styles, particularly those reliant on narrative context and cultural nuances.

%% file: latex/method.tex
In this section, we outline the experiment settings for evaluating LMMs' ability to comprehend humor in comics using the \textsf{PixelHumor} dataset. 

\subsection{Task Definitions}
\label{sec:tasks}

We define four core tasks—\textit{humor identification}, \textit{humor classification}, \textit{humor interpretation}, and \textit{sequence recognition}—to comprehensively evaluate the multimodal humor comprehension capabilities of LMMs. These tasks target key dimensions of cognitive and narrative understanding, from basic cue detection to higher-order reasoning.

\paragraph{Humor Identification.}  
This task tests whether a model can detect and localize humor within a comic. Subtasks include identifying the presence of humor, recognizing contributing factors such as sound effects, determining the most critical panel for humor delivery, and assessing the relative importance of textual versus visual elements. Success indicates basic multimodal perception skills, while failure highlights limitations in recognizing fundamental humor cues.

\paragraph{Humor Classification.}  
Models must categorize comics into one or more humor styles based on a predefined taxonomy (Table~\ref{tab:humorstyles}). As a multi-class classification problem, this task challenges models to disentangle overlapping and subtle humor types. Accurate classification demonstrates an ability to generalize and apply nuanced conceptual understanding—critical given humor’s inherently subjective and context-dependent nature.

\paragraph{Humor Interpretation.}  
This open-ended task assesses a model’s ability to explain why a comic is humorous. Models must generate natural language explanations articulating how textual and visual elements interact to produce comedic effects. This task probes reasoning and abstraction capabilities beyond surface-level pattern recognition, aiming to reveal whether models can ``\textit{think}'' about humor in human-like terms.

\paragraph{Sequence Recognition.}  
Comics rely on a carefully structured narrative order to build context and deliver punchlines. In this task, models must correctly reconstruct the intended sequence of panels and associated textual elements. Success reflects an understanding of temporal dependencies and narrative flow across modalities—an essential aspect of coherent humor comprehension.

Task-specific prompts were designed to align closely with the objectives of each evaluation setting. Full details of the prompt construction are provided in Appendix~\ref{apx:prompts}.

\subsection{Benchmark Models}
\label{sec:benchmarkmodels}
To evaluate LMMs' ability to comprehend humor in comics, we benchmarked a diverse set of models spanning different architectures, parameter scales, and accessibility settings. All models were evaluated under a unified framework using standardized prompts (Appendix~\ref{apx:prompts}) with a temperature of 0 to ensure consistency and deterministic outputs.

We included two closed-source models: GPT-4o\footnote{Evaluated using the GPT-4o 2024-08-06 model}~\cite{achiam2023gpt} and Gemini-1.5-Pro\footnote{Evaluated using the gemini-1.5-pro-001 model}~\cite{reid2024gemini}. GPT-4o represents the latest state-of-the-art in multimodal reasoning, while Gemini-1.5-Pro is a competitive model known for its strong performance in vision-language tasks. 

For open-source evaluation, we selected four models to capture a range of capacities. Among larger models, we tested Qwen2-VL-72B~\cite{Qwen2VL}, a high-capacity vision-language model with strong generalization across multimodal benchmarks, and Gemma3-27B~\cite{team2025gemma}, noted for its advanced visual grounding and contextual reasoning abilities. To complement these, we included two smaller models: LLaVA-OneVision-7B-SI\footnote{The SI variant is pre-trained on single-image tasks, making it suitable for comic panel evaluation}~\cite{li2024llava}, which is optimized for visual understanding at the single-image level, and Qwen2-VL-7B~\cite{Qwen2VL}, a lightweight yet competitive model.

All models were evaluated in their publicly available pre-trained forms without task-specific fine-tuning. This zero-shot setting ensures a fair assessment of their generalization capabilities to novel, humor-centered multimodal reasoning tasks.

%\subsection{Benchmark Models}
%To evaluate LMMs' ability to comprehend humor in comics, we benchmarked two closed-source and two open-source models, ensuring a balanced comparison across diverse architectures. All models were tested using the same prompts (Appendix~\ref{apx:prompts}) with a temperature of 0 to maintain consistency and deterministic outputs. The closed-source models included GPT-4o\footnote{Evaluated using the GPT-4o 2024-08-06 model}~\cite{achiam2023gpt} and Gemini-1.5-Pro\footnote{Evaluated using the gemini-1.5-pro-001 model}~\cite{reid2024gemini}, both known for their strong performance in multimodal tasks. For open-source models, we selected Qwen2-VL-72B~\cite{Qwen2VL} and Gemma3-27B\cite{team2025gemma} as larger models with LLaVA-OneVision-7B-SI\footnote{The SI variant is pre-trained on single-image tasks, making it suitable for our use case.}~\cite{li2024llava} and Qwen2-VL-7B~\cite{Qwen2VL} as smaller models. These models are recognized for their competitive multimodal performance. All models were tested in their pre-trained form without fine-tuning to assess their zero-shot generalization to humor comprehension.

\begin{table}[t]
    \centering
    \small
    \begin{tabular}{p{1.5cm}|c|c|c|c}
    \hline
    \textbf{Sub-Tasks} & \textbf{Model} & \textbf{F1} & \textbf{Prec.} & \textbf{Rec.} \\ \hline\hline
    \multirow{6}{1.5cm}{Humor Presence Identification}  
    & GPT-4o         & 0.983 & 0.983 & 0.988 \\
    & Gemini-1.5-Pro & \textbf{0.984} & 0.982 & 0.988  \\
    & Qwen2-VL-72B   & \textbf{0.984} & 0.983 & 0.988 \\
    & Gemma3 27B     & \textbf{0.984} & \textbf{0.985} & \textbf{0.989} \\
    & LLaVA-OV 7B    & 0.982 & 0.977 & 0.988  \\
    & Qwen2-VL 7B    & 0.982 & 0.979 & 0.987  \\ \hline  
    \multirow{6}{1.5cm}{Sound Effect Identification} 
    & GPT-4o         & \textbf{0.821} & 0.821 & \textbf{0.820}  \\
    & Gemini-1.5-Pro & 0.790 & 0.845 & 0.742 \\
    & Qwen2-VL-72B   & 0.717 & 0.844 & 0.624 \\
    & Gemma3 27B     & 0.743 & 0.840 & 0.666 \\
    & LLaVA-OV 7B    & 0.703 & 0.833 & 0.609 \\
    & Qwen2-VL 7B    & 0.013 & \textbf{0.849} & 0.007 \\ \hline
    \multirow{6}{1.5cm}{Panel Contribution}          
    & GPT-4o         & \textbf{0.765} & \textbf{0.788} & \textbf{0.762} \\
    & Gemini-1.5-Pro & 0.717 & 0.746 & 0.713 \\
    & Qwen2-VL-72B   & 0.494 & 0.605 & 0.485 \\
    & Gemma3 27B     & 0.540 & 0.633 & 0.530 \\
    & LLaVA-OV 7B    & 0.501 & 0.618 & 0.507 \\
    & Qwen2-VL 7B    & 0.517 & 0.639 & 0.498 \\ \hline
    \multirow{6}{1.5cm}{Modality Contribution}       
    & GPT-4o         & \textbf{0.626} & \textbf{0.699} & \textbf{0.656} \\
    & Gemini-1.5-Pro & 0.613 & 0.661 & 0.631 \\
    & Qwen2-VL-72B   & 0.577 & 0.682 & 0.605 \\
    & Gemma3 27B     & 0.211 & 0.696 & 0.352 \\
    & LLaVA-OV 7B    & 0.562 & 0.597 & 0.586 \\
    & Qwen2-VL 7B    & 0.214 & 0.608 & 0.297 \\ \hline
    \end{tabular}
    \caption{Experiment result of various humor identification tasks. The best results are \textbf{bold}.}
    \label{tab:humor_identification_results}
\end{table}

\begin{table*}[t]
    \centering
    \small
    \begin{tabular}{l|c|c|c||c|c|c|c|c|c|c|c|c}
    \hline
    \textbf{Model} & \textbf{F1.} & \textbf{Prec.} & \textbf{Rec.} & \textbf{Com.} & \textbf{Per.} & \textbf{Exa.} & \textbf{Pun.} & \textbf{Sar.} & \textbf{Sil.} & \textbf{Sur.} & \textbf{Dar.} & \textbf{N/A}\\ \hline\hline
    GPT-4o  & \textbf{0.499} & 0.393 & 0.711 & \textbf{0.596} & \textbf{0.965} & \textbf{0.758} & \textbf{0.587} & 0.569 & 0.593 & 0.713 & 0.746 & 0.030   \\  
    Gemini-1.5-Pro & 0.480 & 0.393 & 0.673 & 0.421 & 0.819 & 0.481 & 0.471 & \textbf{0.854} & 0.655 & 0.789 & 0.584 & 0.091  \\ 
    Qwen2-VL 72B & 0.375 & 0.455 & 0.382 & 0.304 & 0.840 & 0.521 & 0.409 & 0.251 & 0.267 & 0.128 & 0.358 & \textbf{0.182}  \\ 
    Gemma3 27B & 0.465 & 0.370 & \textbf{0.735} & 0.446 & 0.767 & 0.576 & 0.333 & 0.689 & \textbf{0.916} & \textbf{0.863} & \textbf{0.827} & 0.091  \\ 
    LLaVA-OV 7B & 0.094 & 0.306 & 0.123 & 0.388 & 0.071 & 0.170 & 0.076 & 0.775 & 0.005 & 0.000 & 0.000 & 0.000  \\ 
    Qwen2-VL 7B & 0.248 & \textbf{0.507} & 0.295 & 0.467 & 0.877 & 0.463 & 0.271 & 0.101 & 0.054 & 0.001 & 0.237 & 0.030  \\ \hline 
    \end{tabular}
    \caption{Experiment result of humor classification task. We also report the Recall score of various humor style types.  \textbf{Com.}: \textit{Comparison}, \textbf{Per.}: \textit{Personification},  \textbf{Exa.}: \textit{Exaggeration}, \textbf{Pun.}: \textit{Pun}, \textbf{Sar.}: \textit{Sarcasm}, \textbf{Sil.}: \textit{Silliness}, \textbf{Sur.}: \textit{Surprise}, \textbf{Dar.}: \textit{Dark}, \textbf{N/A}: \textit{Not Applicable}. The best results are \textbf{bold}.}
    \label{tab:humor_classification_results}
\end{table*}

\subsection{Evaluation Framework and Metrics}
\label{sec:evaluation}
Automated metrics were employed for humor identification, humor classification, and sequence recognition, while human evaluations are conducted for humor interpretation.

For humor identification and humor classification, we used precision, recall, and weighted F1-score against human-annotated ground truth, capturing the models' ability to detect humor and distinguish nuanced humor styles (Table~\ref{tab:humorstyles}). 

Humor interpretation, being open-ended, was evaluated through human ratings on a 7-point Likert scale, assessing the relevance and coherence of the model-generated explanations.

Sequence recognition was evaluated by measuring models’ ability to reconstruct the correct order of panels and associated text. Visual sequencing was assessed using accuracy, while textual sequencing was further evaluated with Word Error Rate (WER) and Character Error Rate (CER) to capture fine-grained alignment.

%% file: latex/experiment.tex
In this section, we present the experimental results across the four humor evaluation tasks: \textit{humor identification}, \textit{humor classification}, \textit{humor interpretation}, and \textit{sequence recognition}. Overall, closed-source models such as GPT-4o and Gemini-1.5-Pro consistently outperform open-source counterparts, particularly on tasks requiring fine-grained narrative understanding and multimodal reasoning. While all models excel at detecting the presence of humor, they struggle significantly with identifying key narrative moments, explaining humor coherently, and reconstructing comic sequences. These findings highlight fundamental challenges in current LMMs' abilities to integrate visual and textual information over extended contexts. We report both quantitative metrics and qualitative analyses to provide a comprehensive assessment of model performance and limitations.

\subsection{Humor Identification}
\label{sec:humor_identification}
Humor identification evaluates a model’s ability to detect humor cues, attribute their sources, and understand the relative contributions of textual and visual elements. Table~\ref{tab:humor_identification_results} summarizes the results across four sub-tasks.

\paragraph{Humor Presence.}
All models achieved near-perfect performance, with F1-scores exceeding 0.98. Gemini-1.5-Pro, Qwen2-VL-72B, and Gemma3-27B recorded the highest F1-scores (0.984), each misclassifying only eight comics. However, given that the dataset is heavily skewed toward humorous comics (only 33 labeled as non-humorous), this high accuracy likely reflects label imbalance rather than deep humor understanding.

\paragraph{Sound Effect Identification.}
While GPT-4o achieved the highest F1-score (0.821) in recognizing the role of sound effects in humor, Qwen2-VL-7B performed poorly. The latter exhibited a strong bias toward predicting sound effects regardless of context, leading to extremely low recall (0.007). These results suggest that some models rely on superficial textual heuristics rather than genuine multimodal understanding.

\paragraph{Panel Contribution.}
Identifying the panel most responsible for humor posed a greater challenge. GPT-4o achieved the highest F1-score, while open-source models lagged significantly. Common errors included defaulting to reading order (e.g., left-to-right selection) rather than recognizing punchline positioning. Notably, Qwen2-VL-72B frequently generated out-of-bounds panel indices, further reducing its performance. These patterns indicate that LMMs struggle to localize key narrative moments crucial for humor delivery.

\paragraph{Modality Contribution.}
This sub-task assessed whether humor was driven by text, visuals, or both. GPT-4o again outperformed others (F1 = 0.626), closely followed by Gemini-1.5-Pro (F1 = 0.613). In contrast, Qwen2-VL-7B and Gemma3-27B achieved low F1-scores, largely due to an over-reliance on predicting ``\textit{Both}'' modalities regardless of context. These findings suggest that fine-grained attribution of humor sources remains a key weakness for many LMMs, particularly smaller or open-source models.

\begin{figure}[t]
    \centering
    \includegraphics[width=1\linewidth]{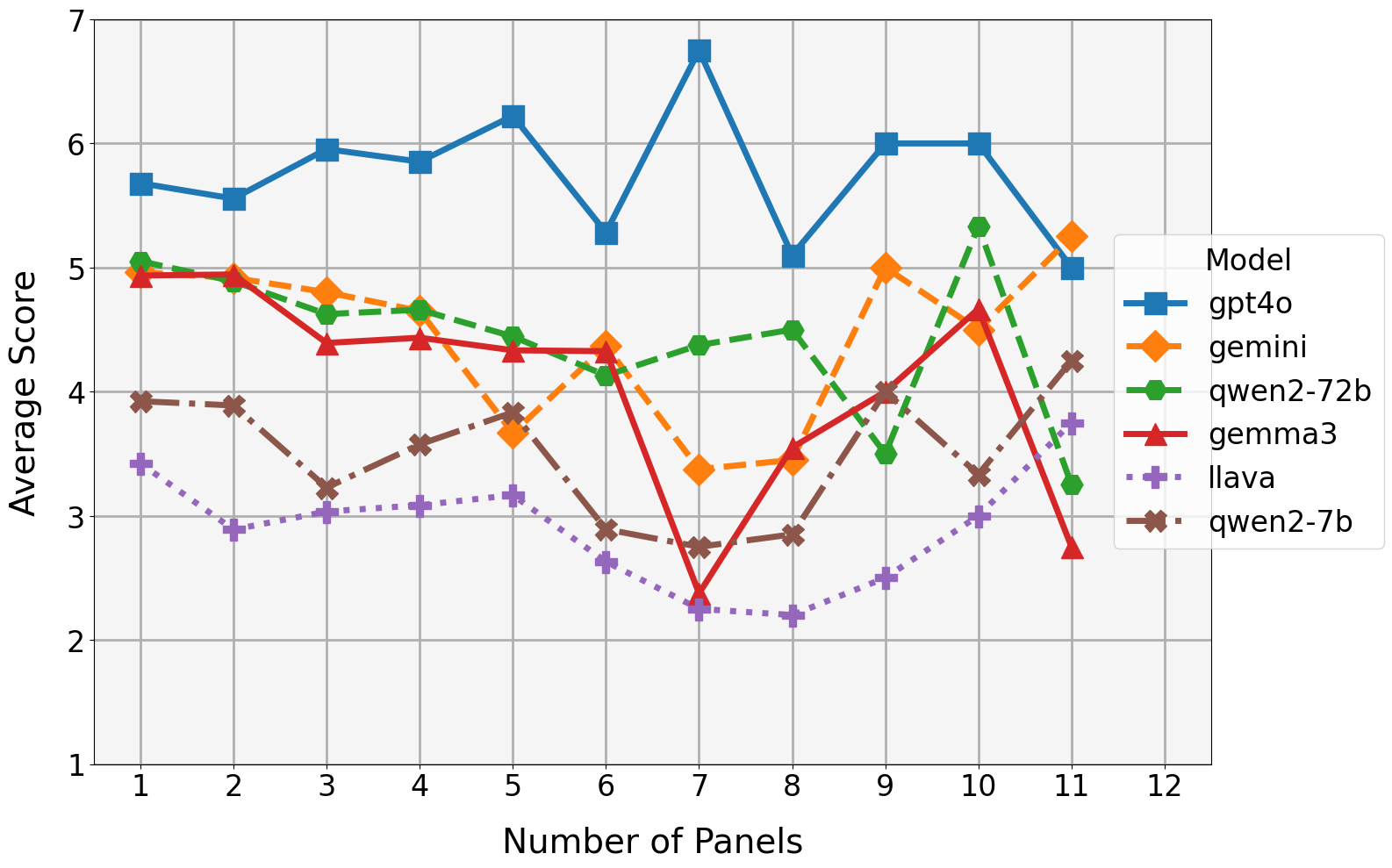}
    \caption{Average relevance scores for the comics (y-axis) plotted against the number of panels (x-axis). Comics with more than 8 panels make up a very small proportion (1.7\%) of the rated comics.}
    \label{fig:line_plot}
\end{figure}

\subsection{Humor Classification}
\label{sec:humor_classification}
Humor classification evaluates a model’s ability to distinguish between different humor styles within a comic. Table~\ref{tab:humor_classification_results} presents the weighted average performance across models.

GPT-4o achieved the highest F1-score, followed closely by Gemini-1.5-Pro and Gemma3-27B. Larger open-source models, such as Qwen2-VL-72B, showed competitive performance but still trailed behind the closed-source models, while smaller models like Qwen2-VL-7B and LLaVA-OV-7B struggled significantly.

% \begin{table}[t]
%     \centering
%     \small
%     \begin{tabular}{l|c|c|c|c|c}
%     \hline
%         \textbf{Model} & \textbf{1 HS} & \textbf{2 HS} & \textbf{3 HS} & \textbf{4 HS} & \textbf{5+ HS}  \\ \hline \hline
%         Ground Truth   & 2,052 & 643 & 96 & 9 & 0 \\
%         GPT-4o         & 343   & 856 & 1,480 & 112 & 9 \\
%         Gemini-1.5-Pro & 39    & 1,227 & 1,239 & 276 & 19 \\
%         Qwen2-VL 72B   & 2,420 & 252 & 119 & 8 & 1 \\
%         Gemma3 27B     & 8     & 140 & 2,226 & 417 & 9 \\
%         LLaVA-OV 7B    & 2,800 & 0 & 0 & 0 & 0 \\
%         Qwen2-VL 7B    & 2,661 & 41 & 50 & 23 & 25 \\ \hline
%     \end{tabular}
%     \caption{Number of humor styles (HS) predicted by each model.}
%     \label{tab:humor_style_num}
% \end{table}

\begin{table}[t]
    \centering
    \small
    \setlength{\tabcolsep}{4pt} % reduce horizontal padding
    \renewcommand{\arraystretch}{0.95} % reduce vertical padding
    \begin{tabular}{l|c|c|c|c|c}
    \hline
        \textbf{Model} & \textbf{1 HS} & \textbf{2 HS} & \textbf{3 HS} & \textbf{4 HS} & \textbf{5+ HS}  \\ \hline \hline
        Ground Truth   & 2,052 & 643 & 96 & 9 & 0 \\
        GPT-4o         & 343   & 856 & 1,480 & 112 & 9 \\
        Gemini-1.5-Pro & 39    & 1,227 & 1,239 & 276 & 19 \\
        Qwen2-VL 72B   & 2,420 & 252 & 119 & 8 & 1 \\
        Gemma3 27B     & 8     & 140 & 2,226 & 417 & 9 \\
        LLaVA-OV 7B    & 2,800 & 0 & 0 & 0 & 0 \\
        Qwen2-VL 7B    & 2,661 & 41 & 50 & 23 & 25 \\ \hline
    \end{tabular}
    \caption{Number of humor styles (HS) predicted by each model.}
    \label{tab:humor_style_num}
\end{table}

A key challenge for open-source models was a strong bias toward assigning a single humor style per comic, despite the multi-label nature of the task (Table~\ref{tab:humor_style_num}). Qwen2-VL-7B, for instance, achieved relatively high precision but low recall, leading to lower F1-score. LLaVA-OV-7B consistently predicted only one humor style (primarily \textit{Sarcasm}), severely limiting its ability to capture the interplay of multiple comedic elements. Similarly, Qwen2-VL-72B and Qwen2-VL-7B exhibited a bias toward predicting \textit{Personification} and \textit{Exaggeration}, reflecting a lack of nuanced humor modeling.

Across all models, \textit{Personification} was classified most accurately, likely due to its clear visual and textual cues, particularly in comics featuring anthropomorphized characters (e.g., \textit{Peanuts}, \textit{Garfield}, \textit{They Can Talk}). In contrast, styles requiring deeper contextual understanding, such as \textit{Sarcasm} and \textit{Dark} humor, were more frequently misclassified, highlighting limitations in models' subtle inference abilities.

Another consistent pattern was poor recall for the ``\textit{N/A}'' (non-humorous) category. This reflects an underlying bias toward humor predictions, likely influenced by the dataset’s skew toward humorous comics. As a result, models struggled to identify non-humorous instances, mirroring trends observed in the humor presence identification task.

\subsection{Humor Interpretation}
\label{sec:humor_interpretation}
We selected 350 comics with high annotator agreement and rated model-generated explanations on a 7-point Likert scale to assess models' ability to generate coherent explanations for comic-based humor. Results are summarized in Table~\ref{tab:humor_interpretation_results}.

\paragraph{Quantitative Analysis.} GPT-4o achieved the highest mean score for humor interpretation, substantially outperforming all open-source models. Gemini-1.5-Pro, Qwen2-VL-72B, and Gemma3-27B followed with comparable mean ratings, while smaller models such as Qwen2-VL-7B and LLaVA-OV-7B performed significantly worse. These results highlight a clear gap between proprietary and open-source models in humor reasoning, as well as the advantage of larger model sizes.

\begin{table}[t]
    \centering
    \small
    \begin{tabular}{l|c|c|c}
    \hline
    \textbf{Model} & \textbf{Mean} & \textbf{Median} & \textbf{STD} \\ \hline \hline
    GPT-4o         & \textbf{5.801} & \textbf{6.500} & 1.371 \\
    Gemini-1.5-Pro & 4.647 & 5.000 & 1.687 \\
    Qwen2-VL 72B   & 4.653 & 4.500 & 1.437 \\
    Gemma3 27B     & 4.439 & 4.500 & 1.796 \\
    LLaVA-OV 7B    & 3.039 & 3.000 & 1.595 \\
    Qwen2-VL 7B    & 3.477 & 3.500 & 1.712 \\ \hline
    \end{tabular}
    \caption{Statistics of relevance scores evaluated by human annotators on LMMs' generated interpretation of the comics. The best results are \textbf{bold}.}
    \label{tab:humor_interpretation_results}
\end{table}

% \begin{table}[t]
%     \centering
%     \small
%     \begin{tabular}{l|c|>{\centering\arraybackslash}p{1cm}|c|c|c}
%     \hline
%     \textbf{Model} & \textbf{K-alpha} & \textbf{Exact Match} & \textbf{Mean} & \textbf{Median} & \textbf{STD} \\ \hline \hline
%     GPT-4o         & 0.252 & 0.368 & 1.334 & 1.000 & 1.534 \\
%     Gemini-1.5-Pro & 0.367 & 0.257 & 1.694 & 1.000 & 1.553 \\
%     Qwen2-VL 72B   & 0.391 & 0.265 & 1.289 & 1.000 & 1.154 \\
%     Gemma3 27B     & 0.476 & 0.243 & 1.251 & 1.000 & 1.069 \\
%     LLaVA-OV 7B    & 0.456 & 0.237 & 1.551 & 1.000 & 1.394 \\
%     Qwen2-VL 7B    & 0.444 & 0.274 & 1.491 & 1.000 & 1.379 \\ \hline
%     Overall        & 0.523 & 0.274 & 1.435 & 1.000 & 1.368 \\ \hline
%     \end{tabular}
%     \caption{Inter-rater reliability score and statistics of absolute rating difference broken down by models.}
%     \label{tab:irr_humor_interpret}
% \end{table}

\begin{table}[t]
    \centering
    \small
    \setlength{\tabcolsep}{3pt} % default is 6pt, reduce horizontal padding
    \begin{tabular}{l|c|>{\centering\arraybackslash}p{0.9cm}|c|c|c}
    \hline
    \textbf{Model} & \textbf{K-alpha} & \textbf{Exact Match} & \textbf{Mean} & \textbf{Median} & \textbf{STD} \\ \hline \hline
    GPT-4o         & 0.252 & 0.368 & 1.334 & 1.000 & 1.534 \\
    Gemini-1.5-Pro & 0.367 & 0.257 & 1.694 & 1.000 & 1.553 \\
    Qwen2-VL 72B   & 0.391 & 0.265 & 1.289 & 1.000 & 1.154 \\
    Gemma3 27B     & 0.476 & 0.243 & 1.251 & 1.000 & 1.069 \\
    LLaVA-OV 7B    & 0.456 & 0.237 & 1.551 & 1.000 & 1.394 \\
    Qwen2-VL 7B    & 0.444 & 0.274 & 1.491 & 1.000 & 1.379 \\ \hline
    Overall        & 0.523 & 0.274 & 1.435 & 1.000 & 1.368 \\ \hline
    \end{tabular}
    \caption{Inter-rater reliability and statistics of absolute rating difference by model.}
    \label{tab:irr_humor_interpret}
\end{table}

Figure~\ref{fig:line_plot} illustrates how comic length affects relevance scores. Performance remains relatively stable for comics with four or fewer panels, but begins to fluctuate as narratives grow longer, particularly beyond six panels. This trend suggests that while additional context can sometimes aid humor interpretation, longer sequential dependencies pose increasing challenges for LMMs.

\paragraph{Qualitative Analysis.} Generated interpretations from smaller open-source models frequently hallucinated humor elements or produced generic, template-like explanations (See Appendix~\ref{apx:human_explanation_examples} for examples). While larger open-source models provided more plausible interpretations, they too struggled as the number of panels increased. Even top-performing closed-source models exhibited volatility in longer narratives, indicating persistent difficulties in maintaining multimodal coherence over extended sequences.

These findings point to long-context modeling as a critical bottleneck for humor comprehension. Future improvements may benefit from hierarchical attention mechanisms, enhanced cross-modal alignment, or human-in-the-loop training strategies to better support narrative tracking and sequential humor reasoning.

\paragraph{Inter-evaluator Agreement.} After evaluating the generated explanations for the humor interpretation task, we calculated the inter-rater reliability shown in Table \ref{tab:irr_humor_interpret}. The Krippendorff’s alpha is 0.523 which reflects a moderate agreement between evaluators. This value is within the acceptable range for subjective tasks in NLP and HCI, such as emotion or empathy evaluation. Moreover, the low median rating difference (1.0) and standard deviation ($\sim$ 1.4) show that even when evaluators disagree, their ratings are generally close in magnitude. GPT-4o received the highest exact match agreement but a lower K-alpha as it's ratings skew toward the upper end (scores 6–7); these higher scores reduces the score variance and suppresses alpha despite consistent judgments, a known behavior of this metric. These values indicate that the human evaluations of the LMM's explanations remain largely consistent, despite the subjective nature of humor interpretation.

\subsection{Human Preferences.}
We further assessed the difficulty of humor interpretation for LMMs through a preference study. In this study, we selected 70 comics: 10 comics randomly chosen per source and a human was tasked to interpret the humor in these comics. After which, two participants were tasked to choose the best interpretation of the humor---one human and six generated---for the comics. Disagreements are resolved through a third annotator from the chosen explanations.

Human written explanations are mostly preferred over generated ones in interpreting humor. In 48 (68.6\%) of the 70 samples (Table~\ref{tab:humor_explanation_choice}), the human written explanation was judged to be the best by human participants, despite the high mean scores achieved by GPT-4o in this task (Table~\ref{tab:humor_interpretation_results}). We attribute this outcome to two main factors: the superior reasoning abilities of humans and the repeated general reasoning patterns for humor by the models. Human-written explanations were able to better combine information from both modalities and describe how they synergize to amplify humor. In contrast, generated explanations often give general explanations that rely on absurdity or unexpectedness, which might not be the main crux of the humor (see examples in Appendix~\ref{apx:human_explanation_examples}). Ultimately, these results indicate that humans are still far ahead of LMMs in more complex multimodal humor tasks: fine-grained comprehension and interpretation.

\begin{table}
    \centering
    \small
    \begin{tabular}{l|c}
    \hline
    \textbf{Explanation Source} & \textbf{Times Selected}\\ \hline \hline
    Human          & 48 \\
    GPT-4o         & 14 \\
    Gemini-1.5-Pro & 3  \\
    Qwen2-VL 72B   & 3  \\
    Gemma3 27B     & 1  \\
    LLaVA-OV 7B    & 0  \\
    Qwen2-VL 7B    & 1  \\ \hline
    \end{tabular}
    \caption{Results of the human preference study to select the best explanation out of the six generated and one human written explanation for 70 comics.}
    \label{tab:humor_explanation_choice}
\end{table}

\subsection{Sequence Recognition}
\label{sec:sequence_recognition}
Sequence recognition---visual and text---evaluates a model’s ability to reconstruct the correct reading order of comic panels, a crucial component for maintaining narrative coherence and delivering humor. Each panel was randomly assigned a unique numerical label from 1 to \(N\), displayed at the top-left corner.

\paragraph{Visual Sequence.} Visual sequence recognition assesses a model's ability to reconstruct the correct order of panels in the comics. Gemini-1.5-Pro achieved the highest panel sequencing accuracy (0.645), closely followed by GPT-4o (0.614), while open-source models lagged significantly (e.g., Qwen2-VL-72B at 0.343, Gemma3-27B at 0.310) (Table~\ref{tab:sequence_recognition_results}). LLaVA-OV-7B showed intermediate performance (0.422), outperforming other open models but still far behind the closed-source models.

Qualitative analysis for the visual sequencing task revealed several common failure patterns across all models. Many defaulted to conventional reading orders (e.g., left-to-right, top-to-bottom) without adapting to the randomized panel numbering. Others miscounted the number of panels or produced continuous sequences until reaching output limits, suggesting reliance on heuristic completions rather than true sequential reasoning. These errors underscore broader limitations in LMMs’ capacity for structured visual narrative processing, a key requirement for coherent humor understanding.

%\subsection{Sequence Recognition}
%\paragraph{Panel Sequence Recognition.}  
%This task evaluates the model's ability to determine the correct reading order of comic panels, which is essential for understanding narrative progression and humor delivery. Each comic's panels were randomly labeled from 1 to \(N\) (maximum number of panels) without repetition, with numerical indicators placed at the top left corner. 

%Among the models, Gemini-1.5-Pro achieved the highest accuracy (0.645), followed closely by GPT-4o (0.614), while the open-source models Qwen2-VL 72B (0.343), Gemma3 27B (0.310), LLaVA-OV 7B (0.422) and Qwen2-VL 7B (0.314) performed significantly worse (Table~\ref{tab:sequence_recognition_results}). 

%Examining the error cases, we noted that all models exhibited common failure patterns. Many relied on conventional reading orders (e.g., left-to-right, top-to-bottom), rather than correctly identifying the intended panel sequence. Some miscounted the total number of panels, generating nonexistent indices, while others produced continuous number sequences until reaching the output token limit, indicating heuristic completions rather than genuine reasoning. These issues highlight LMMs’ broader limitation in processing sequential visual narratives, a crucial skill for interpreting humor based on narrative structure. 

    \begin{table}[t]
        \centering
        \small
        \begin{tabular}{c|p{1cm}||p{1cm}|c|c}
        \hline
        \textbf{Model} & \textbf{Panel Acc.}& \textbf{Text Acc.} & \textbf{WER} & \textbf{CER}\\ \hline \hline
        GPT-4o         & 0.614 & \textbf{0.326} & \textbf{0.230} & \textbf{0.241} \\
        Gemini-1.5-Pro & \textbf{0.645} & 0.258 & 0.328 & 0.339  \\
        Qwen2-VL 72B   & 0.343 & 0.176 & 0.443 & 0.496  \\
        Gemma3 27B     & 0.310 & 0.141 & 0.366 & 0.419  \\
        LLaVA-OV 7B    & 0.422 & 0.130 & 0.415 & 0.515  \\
        Qwen2-VL 7B    & 0.314 & 0.103 & 0.463 & 0.485  \\ \hline
        \end{tabular}
        \caption{Experiment results for sequence recognition tasks. The best results are \textbf{bold}.}
        \label{tab:sequence_recognition_results}
    \end{table}

\paragraph{Text Sequence.} Text sequence recognition assesses a model’s ability to reconstruct the correct order of dialogue or captions within comics. Table~\ref{tab:sequence_recognition_results} summarizes the results. GPT-4o achieved the highest accuracy (0.326) with the lowest error rates (WER = 0.230, CER = 0.241), demonstrating strong alignment between text elements and narrative structure. Gemini-1.5-Pro followed with slightly lower performance (Text Acc. = 0.258, WER = 0.328, CER = 0.339), while all open-source models lagged significantly. Qwen2-VL-7B performed worst (Text Acc. = 0.103, WER = 0.463, CER = 0.485).

Qualitative analysis for text sequencing revealed that larger open-source models, such as Qwen2-VL-72B and Gemma3-27B, occasionally produced repetitive or degenerate outputs (e.g., ``\textit{Ctrl-C, Ctrl-V...}''), highlighting weaknesses in maintaining coherent narrative flow despite increased model capacity. Smaller models frequently hallucinated or omitted key text elements, further disrupting generation.

These findings underscore that even advanced LMMs face persistent challenges in structured multimodal reasoning. Bridging the gap between panel and text sequencing performance will require training strategies that more effectively model sequential dependencies across modalities.

%% file: latex/discussion.tex
This work introduced \textsf{PixelHumor}, a novel benchmark designed to evaluate LMMs ability to understand humor in online comics. Through comprehensive experiments across four core tasks, humor identification, humor classification, humor interpretation, and sequence recognition, we provided the first in-depth assessment of LMMs' multimodal humor comprehension.

Our results reveal that while LMMs achieve near-perfect accuracy in detecting the presence of humor, they struggle with deeper aspects such as narrative sequencing, subtle humor style classification, and modality attribution. Sequence recognition results show that even top models often rely on conventional reading heuristics rather than accurately reconstructing narrative flow. In humor classification, models consistently excel at explicit styles like \textit{personification} but perform poorly on nuanced categories such as \textit{sarcasm} and \textit{dark humor}. Additionally, humor interpretation tasks highlight significant degradation in reasoning quality as narrative complexity increases.

These findings have important implications for the development of future multimodal AI systems. Our results suggest that current LMMs rely heavily on surface-level heuristics---such as favoring the textual modality---rather than engaging in true multimodal contextual and sequential reasoning. To genuinely comprehend humor, models must move beyond pattern recognition to understand causal, temporal, and multimodal relationships that underpin comic narratives. This will require advances such as hierarchical modeling of narrative structures, enabling models to track setups, twists, and punchlines across multiple panels and modalities. Moreover, improved cross-modal fusion mechanisms are needed to better integrate visual and textual cues, particularly for humor that emerges from their interaction rather than from either modality alone. %Cultural awareness also plays a crucial role, as humor is often culturally dependent, and LMMs trained predominantly on English-centric or Western data may fail to capture diverse humor expressions. %Ultimately, scaling model size alone is insufficient; genuine progress in humor understanding will require architectural innovations and more nuanced training approaches.

%These findings have important implications: they suggest that current LMMs rely heavily on surface-level heuristics rather than true contextual and sequential reasoning. Improving humor understanding will thus require advances beyond scaling model size—specifically, in areas such as hierarchical modeling of narrative structures, enhanced cross-modal reasoning, and culturally-aware humor comprehension.

Future work could focus on developing models that better track long-range multimodal dependencies, distinguish finer-grained humor mechanisms, and adapt to diverse cultural humor norms. Extending \textsf{PixelHumor} to cover broader linguistic and stylistic variations could further stress-test LMMs' reasoning limits. Future work could also cover web comics in other languages to assess an LMM's ability to comprehend humor in multilingual settings.

%% file: latex/conclusion.tex
\textsf{PixelHumor} is a challenging benchmark, comprised of web comics, to assess an LMM's ability to comprehend and interpret humor. Our experiments highlight fundamental limitations in LMMs, and show that they are still far behind humans in complex humor-related tasks: in particular, humor interpretation where human written explanations are still largely preferred; humor style classification where models struggle to identify darker and sarcastic styles; and sequence recognition which is required in the build-up to a humorous punchline. This work lays the foundation for building AI systems that can engage more socio-intelligently with one of the most complex forms of human communication: humor.

%% file: latex/limitations.tex
While PixelHumor marks a significant advancement in evaluating humor comprehension in LMMs, several limitations remain. Humor is inherently subjective, and despite rigorous annotation guidelines, screening, and training, individual biases may influence interpretations. Additionally, PixelHumor focuses on static visual-textual humor in comics, excluding temporal and dynamic elements present in videos and animations, which are critical to humor in other media formats. Furthermore, as the dataset is primarily in English and primarily based in Western media, it may over-represent Western humor conventions, limiting cross-cultural generalizability. Expanding PixelHumor to include multilingual and culturally diverse humor sources would be a valuable direction for future research.

% While PixelHumor marks a significant advancement in evaluating humor comprehension in LMMs, several limitations remain. Humor is inherently subjective, and despite rigorous annotation guidelines, screening, and training, individual biases may influence interpretations. Additionally, PixelHumor focuses on static visual-textual humor in comics, excluding temporal and dynamic elements present in videos and animations, which are critical to humor in other media formats. Furthermore, as the dataset is primarily in English, it may over-represent Western humor conventions, limiting cross-cultural generalizability. Expanding PixelHumor to include multilingual and culturally diverse humor sources would be a valuable direction for future research.

%While PixelHumor represents a significant step towards evaluating humor comprehension in LMMs, it is essential to acknowledge certain limitations. Humor is inherently subjective, and despite our efforts to mitigate biases through detailed annotation guidelines, screening and trainings, individual preferences and interpretations can introduce variations in the annotations. Further, this work primarily focuses on visual and textual modalities in comics, excluding temporal or dynamic elements which are prevalent in other forms of media such as videos or animations. Furthermore, the comics in PixelHumor are primarily in English, potentially over-representing Western humor. Consequently, humor specific to other cultures, groups, or languages may be less represented. 

%% file: latex/ethics.tex
We collected the web comics from a diversified public online sources to capture a broad spectrum of humor styles. However, comics such as Cyanide and Happiness may contain violence, sexually explicit and dark humor that might be uncomfortable for some readers. We are including these comics as they are still valuable for future work in understanding dark humor, hate speech and their connections in intelligent multimodal systems. As a precaution, we will separate these comics in the benchmark and label them as ``potentially harmful''.

All annotators were recruited through our university's internal calls for part-time research assistantship. There were no academic incentives towards the annotators and paid for 20 dollars per hour, which exceeds the minimum wage in our local context (9 dollars per hour). The annotators were informed in advance that the dataset may contain examples of dark or sarcastic humor, and they could withdraw or skip any samples they found uncomfortable. Additionally, annotators were compensated fairly and were provided with clear annotation guidelines that focused on humor classification, not value judgments or endorsements of the content.

% All annotators were informed in advance that the dataset may contain examples of dark or sarcastic humor, and they could withdraw or skip any samples they found uncomfortable. Additionally, annotators were compensated fairly and were provided with clear annotation guidelines that focused on humor classification, not value judgments or endorsements of the content. 

While releasing our dataset, we do not redistribute the images directly, but instead provide URLs and associated metadata. We also plan to include a detailed usage disclaimer and data documentation sheet (datasheet for datasets) that explicitly warns users of potential content sensitivities and encourages ethical use of the dataset in research contexts.

%% file: latex/acknowledgement.tex
This research is supported by the National Research Foundation, Singapore under its AI Singapore Programme (AISG Award No: AISG3-AMP-2024-08-001). We also sincerely thank Bryan Tan (Chen Zhengyu) and Jia Wang Peh for their additional annotations and evaluations, which  substantially improved this work.

%% file: latex/Appendix.tex
\section{Comparison with Existing Datasets}
\label{apx:comparison}

Table~\ref{tab:novelty} shows the detailed difference between PixelHumor and other humor datasets. While we acknowledge the existence of prior humor-related datasets, these resources typically focus on single-panel, single-modality inputs and are limited to narrow task scopes, such as binary humor classification or caption ranking. In contrast, PixelHumor introduces a new paradigm for humor understanding that centers on multimodal, multi-panel narrative reasoning. Our dataset supports four distinct yet complementary tasks: humor detection, multi-label humor classification, explanation via panel contribution, and sequence reconstruction, offering a comprehensive and fine-grained benchmark for evaluating LMMs. To our knowledge, this is the first dataset to jointly evaluate LMMs’ capabilities across both temporal structure and modality attribution in humor, making PixelHumor a uniquely valuable resource for advancing this research area.

    \begin{table*}
        \small
        \centering
        \begin{tabular}{|p{2cm}|p{1.2cm}|p{1.6cm}|p{2cm}|p{4cm}|p{1.5cm}|p{1.5cm}|}
        \hline
        \textbf{Dataset} & \textbf{Modality} & \textbf{Single / Multi panel} & \textbf{Humor Styles} & \textbf{Task Coverage} & \textbf{Sequential Reasoning} & \textbf{Attribution Granularity}    \\ \hline \hline
        \textbf{PixelHumor (ours)} & I + T & Multi-panel & 8 styles (multi-label) & Humor detection, Humor interpretation, Humor classification, Sequence Recogniiton & Yes & Panel + Modality \\ \hline
        New Yorker Caption Contest \cite{hessel-etal-2023-androids} & I + T & Single-panel & N/A & Caption Evaluation (Ranking, Matching) & No & Whole image-text pair \\  \hline
        Memotion 3.0 \cite{mishra2023memotion} & I + T & Multi-panel & Sarcasm & Sentiment analysis, Emotion classification, Scales/Intensity of Emotion Classes & No & Whole image-text pair \\  \hline
        YesBut \cite{nandy2024yesbut} & I + T & Multi-panel & Satire & Satirical Image Detection, Satirical Image Understanding, Satirical Image Completion & No & Whole image-text pair \\  \hline
        HumorDB \cite{jain2024ai} & I & Single-panel & N/A & Detection, Interpretation, Humor Comparison & No & Image \\  \hline
        TalkFunny \cite{chen2024talk} & T & N/A & Affiliative, Self-enhancing, Aggressive, Self-defeating & Humor Sentiment Style Classification, Humor Generation & No & Text-only \\  \hline
        One-liners \cite{mihalcea-strapparava-2005-making} & T & N/A & Alliteration, Antonymy, Adult Slang & Humor detection & No & Text-only \\  \hline
        Pun of the day \cite{yang-etal-2015-humor} & T & N/A & Pun & Humor detection, Humor anchor extratcion & No & Text-only \\  \hline
        Ted-Laughter \cite{chen-lee-2017-predicting} & T & N/A & N/A & Humor detection & No & Text-only \\ \hline
        \end{tabular}
        \caption{Summary of existing humor datasets. T: Text, I: Image, V: Video, A: Audio.}
        \label{tab:novelty}
    \end{table*}

\section{Annotation Guidelines and Questions}
\label{apx:annotation_guidelines}
Annotators answered the following five questions for each comic, with a clear purpose and rationale outlined for each task:

\begin{enumerate}
    \item \textit{Do you understand the humor in this comic? (Y/N)}  
        \par \textbf{Purpose:} To determine whether humor is present in the comic.  
        \par \textbf{Rationale:} Focuses on identifying the intended humor, independent of the annotator’s personal amusement, ensuring objectivity.
    \item \textit{Do the words with sound effects contribute to the humor in this comic? (MCQ: Present, contribute; Present, do not contribute; Absent) Choose Absent if there are no sound effects in this comic.}
        \par \textbf{Purpose:} To assess the contribution of onomatopoeic sound effects to the comic’s humor.  
        \par \textbf{Rationale:} This objective measure serves as a control question for quality checks.
    \item \textit{Type the panel number that contributes the most to the humor. (NA if no humor is present)}  
        \par \textbf{Purpose:} To identify the punchline panel.  
        \par \textbf{Rationale:} Helps pinpoint the comic's critical humorous element, even with randomized panel numbering.
    \item \textit{Do the text or the visuals in this comic contribute more to the humor? (MCQ: Text, Visual, Both - Equal Contribution, NA) Indicate NA if this comic is not humorous to you.}  
        \par \textbf{Purpose:} To assess the relative importance of text and visuals in the comic's humor.  
        \par \textbf{Rationale:} Evaluates multimodal reasoning, where humor may arise from the interaction of textual and visual elements.
    \item \textit{Which humor styles best describe the comic? (Select all that apply or NA)}  
        \par \textbf{Purpose:} To classify the comic into relevant humor styles from the defined taxonomy.  
        \par \textbf{Rationale:} Enables a nuanced categorization of humor styles, ensuring alignment with the dataset's taxonomy.
\end{enumerate}

\pagebreak
\section{Humor Interpretation Guidelines}
\label{apx:humor_interpretation_guidelines}
Evaluators are presented with an interpretation of the humor in the comic. They are to evaluate this explanation by giving it a score from 1 (strongly disagree) to 7 (strongly agree) and state their justifications for their score. The pointers below were provided to help the evaluators rate and write their justifications for the model's explanation:

\begin{itemize}
    \item Give an overall justification or reason for the rating of the comic.
    \item Reference exactly which part of the explanation is accurate or inaccurate with direct quotes.
    \item Elaborate on any points that the explanation may have missed out or are accurate in their inference (e.g. being able to infer the true intentions of the narrative which are commonly found in styles like sarcasm).
    \item Reference the taxonomy of humor styles (Table \ref{tab:humorstyles}) as detailed in Question 5 as needed.
\end{itemize}

The participants \textbf{must fulfill at least two out of four of these pointers} in their justification for their score to be accepted as a valid evaluation. Below, we provided examples of accepted or rejected justifications.

\newtcolorbox[auto counter, number within=section]{mutatorbox1}[2][]{colback=gray!10!white, colframe=black, coltitle=black, title=Mutator Prompts~\thetcbcounter: #2,#1, sharp corners, boxrule=0.5mm, left=10mm, right=10mm, fonttitle=\bfseries, halign=left, valign=left}
  
\begin{tcolorbox}[colframe=gray!40!black, colback=gray!5!white, title=Example Justification for Humor Interpretations]
{\small
\textbf{Example Explanation:}
"The comic's humor stems from the unexpected juxtaposition of traditional religion and modern technology. We see a priest soliciting a donation, a common practice, but the punchline subverts our expectations. Instead of offering a typical excuse, the person being solicited implies they can donate digitally through an app called "Papal," a playful reference to the Pope. This unexpected twist, highlighting the increasing digitization of even traditionally cash-based interactions like charitable giving, creates the humor."

\vspace{0.1cm}

\textbf{Good/Accepted Justification:} The overall explanation is accurate, but the phrase “humor stems from the unexpected juxtaposition of traditional religion” is inaccurate as the main humor style is "surprise". The comic also uses a pun which was not reflected in the explanation. Rating: 5/7

\vspace{0.1cm}

%\textbf{Rejected Justifications:}
%\begin{itemize}
%    \item The overall explanation is accurate. \red{rejected (only fulfill one out of the four pointers provided)}
%    \item The explanation is funny as ... \red{rejected (not an evaluation and does not show the evaluator's rationale for the score)}
%\end{itemize}

}
\end{tcolorbox}

\section{Additional Data Analysis}
\label{apx:dataanalysis}
\begin{table}[H]
    \centering
    \small
    \begin{tabular}{l|c|c|c}
    \hline
    \textbf{Data Source} & \textbf{Max} & \textbf{Min} & \textbf{Average} \\ \hline\hline
    \textit{Happiness and Cyanide}          & 17  & 1  & 4.34 \\
    \textit{Peanuts}          & 11  & 2  & 4.53 \\
    \textit{Garfield}         & 8   & 2  & 3.50 \\
    \textit{XKCD}             & 10  & 1  & 2.57 \\
    \textit{PhD Comics}      & 8   & 1  & 3.67 \\
    \textit{They Can Talk}    & 7   & 1  & 2.87 \\
    \textit{SMBC}             & 18  & 1  & 3.49 \\
    \hline
    \end{tabular}
    \caption{Statistical distribution of panels in \textsf{PixelHumor} dataset breakdown by data source.}
    \label{tab:panel_stats}
\end{table}

\begin{table}[H]
    \centering
    \small
    \begin{tabular}{l|c|c|c}
    \hline
    \textbf{Data Source} & \textbf{Max} & \textbf{Min} & \textbf{Average} \\ \hline\hline
    \textit{Happiness and Cyanide}          & 161  & 0  & 24.87 \\
    \textit{Peanuts}          & 152  & 0  & 38.10 \\
    \textit{Garfield}         & 90   & 1  & 18.66 \\
    \textit{XKCD}             & 253  & 0  & 39.29 \\ 
    \textit{PhD Comics}       & 140  & 0  & 46.71 \\
    \textit{They Can Talk}    & 45   & 0  & 15.27 \\
    \textit{SMBC}             & 501  & 0  & 59.79 \\
    \hline
    \end{tabular}
    \caption{Statistical distribution of words in \textsf{PixelHumor} dataset breakdown by data source.}
    \label{tab:word_stats}
\end{table}

Tables~\ref{tab:panel_stats} and \ref{tab:word_stats} present the statistical distributions of panels and words in the \textsf{PixelHumor} dataset, respectively. The dataset includes highly complex comics, with some featuring up to 18 panels and 501 words, as observed in \textit{SMBC}. On average, comics from \textit{Peanuts} and \textit{Happiness and Cyanide} tend to have a higher number of panels, while those from \textit{PhD Comics} and \textit{SMBC} are notably more text-heavy.

Table~\ref{tab:humor_style_stats} provides a breakdown of the \textsf{PixelHumor} dataset by data source and humor styles, showcasing the diversity of humor types represented across the various sources.

\begin{table*}[t]
    \centering
    \small
    \begin{tabular}{l|c|c|c|c|c|c|c|c|c}
    \hline
    \textbf{Data Source} & \textbf{Com.} & \textbf{Per.} & \textbf{Exa.} & \textbf{Pun.} & \textbf{Sar.} & \textbf{Sil.} & \textbf{Sur.} & \textbf{Dar.} & \textbf{N/A}\\ \hline\hline
    \textit{Happiness and Cyanide} & 13 & 7 & 28 & 95 & 7 & 58 & 145 & 42 & 5  \\ 
    \textit{Peanuts} & 8 & 76 & 92 & 9 & 22 & 87 & 96 & 2 & 8  \\ 
    \textit{Garfield} & 7 & 131 & 43 & 5 & 59 & 85 & 69 & 1 & 0  \\ 
    \textit{XKCD} & 44 & 5 & 46 & 20 & 30 & 130 & 98 & 11 & 16  \\ 
    \textit{PhD Comics} & 59 & 4 & 104 & 20 & 39 & 56 & 116  & 1 & 1 \\ 
    \textit{They Can Talk} & 13 & 336 & 2 & 1 & 4 & 17 & 25 & 1 & 1 \\ 
    \textit{SMBC} & 47 & 14 & 81 & 14 & 13 & 97 & 106 & 26 & 2 \\ \hline 
    \end{tabular}
    \caption{Breakdown of \textsf{PixelHumor} dataset by data source and humor types. \textbf{Com.}: \textit{Comparison}, \textbf{Per.}: \textit{Personification},  \textbf{Exa.}: \textit{Exaggeration}, \textbf{Pun.}: \textit{Pun}, \textbf{Sar.}: \textit{Sarcasm}, \textbf{Sil.}: \textit{Silliness}, \textbf{Sur.}: \textit{Surprise}, \textbf{Dar.}: \textit{Dark}, \textbf{N/A}: \textit{Not Applicable}.}
    \label{tab:humor_style_stats}
\end{table*}

\section{Task-Specific Humor Evaluation Prompts}
\label{apx:prompts}
We designed the following prompts to evaluate LMMs on the four humor evaluation tasks outlined in Section~\ref{sec:tasks}. Questions 1–4 assess humor identification, testing the model's ability to detect and analyze the presence of humor. Question 5 evaluates humor classification, supported by guidelines and definitions of the humor styles. Question 6 focuses on humor interpretation, requiring the model to articulate the reasoning behind the humor. Finally, Questions 7 and 8 address sequence recognition, evaluating the model's ability to understand the intended order of textual and visual elements.

\newtcolorbox[auto counter, number within=section]{mutatorbox2}[2][]{colback=gray!10!white, colframe=black, coltitle=black, title=Mutator Prompts~\thetcbcounter: #2,#1, sharp corners, boxrule=0.5mm, left=10mm, right=10mm, fonttitle=\bfseries, halign=left, valign=left}
  
\begin{tcolorbox}[colframe=gray!40!black, colback=gray!5!white, title=Prompts for Humor Identification Task]
{\small

\textbf{System prompt}: \textit{You are a humorous assistant that understands comics. You will be given comics and your task is to evaluate the comics.}
\\ \\
\textbf{Question 1}: \textit{Do you understand the humor of this comics? Please output only a single word answer ``Yes'' or ``No''.}
\\ \\
\textbf{Question 2}: \textit{Analyze the comic and respond based on the following criteria regarding text-based sound effects:
(a) If there are no sound effects present in the comic, output "Absent".
(b) If sound effects are present and contributing to the humor, output "Present, contribute".
(c) If sound effects are present but do not contribute to the humor, output "Present, do not contribute".}
\\ \\
\textbf{Question 3}: \textit{Which panel contributes the most to the humor of this comic? Please output only the labeled panel number.}
\\ \\
\textbf{Question 4}: \textit{Is the text or the visual modality more important to the humor in this comic? Output "Both" if both modalities contribute humor to the comic. Please output only a single word answer ``Text'', ``Visual'', ``Both''.}
}
\end{tcolorbox}

\newtcolorbox[auto counter, number within=section]{mutatorbox4}[2][]{colback=gray!10!white, colframe=black, coltitle=black, title=Mutator Prompts~\thetcbcounter: #2,#1, sharp corners, boxrule=0.5mm, left=10mm, right=10mm, fonttitle=\bfseries, 
  halign=left, valign=left}
  
\begin{tcolorbox}[colframe=gray!40!black, colback=gray!5!white, title=Prompts for Humor Classification Task]
{\small

\textbf{System prompt}: \textit{You are a humorous assistant that understands comics. You will be given comics and your task is to evaluate the comics.}
\\ 
\textbf{Question 5}: \textit{Which humor styles best describe the comic? Here are some guidelines for each humor style.
\\
\textbf{Comparison}: This comic compares two or more objects/ideas to reference the differences or similarities. This comic is funny because of this comparison.
\\
\textbf{Personification}: This comic has at least one animal/creature/plant that acts like a human (talking, running on two legs etc.). This comic is funny because of this personified creature/plant.
\\
\textbf{Exaggeration}: This comic attempts to exaggerate (overemphasize or magnify) something out of proportion. This comic is funny because of this exaggeration/absurdity.
\\
\textbf{Pun}: This comic is funny because of the linguistic elements. Linguistic elements include: uncommon uses of language, double-meanings in phrases or words etc.
\\
\textbf{Sarcasm}: This comic expresses an idea/thought that is not the real intention of the character/comic. This comic is funny because of the sarcasm present.
\\
\textbf{Silliness}: There are elements in the comic which are absurd and/or ridiculous. The characters are or did something foolish. This comic is funny because of the silly elements.
\\
\textbf{Surprise}: There was a twist in the narrative or an unexpected element in the comic. This comic is funny because of the twist or unexpected elements.
\\
\textbf{Dark}: There are potentially sensitive, taboo or ideas that violate the norm in this comic where if taken out of context in this comic, might be offensive to others. This comic is because of these benign violations or the dark humor present.
\\You may select multiple humor styles but output only the humor styles ``Comparison'', ``Personification'', ``Exaggeration'', ``Pun'', ``Sarcasm'', ``Silliness'', ``Surprise'' or ``Dark''.}
}
\end{tcolorbox}

\newtcolorbox[auto counter, number within=section]{mutatorbox3}[2][]{colback=gray!10!white, colframe=black, coltitle=black, title=Mutator Prompts~\thetcbcounter: #2,#1, sharp corners, boxrule=0.5mm, left=10mm, right=10mm, fonttitle=\bfseries, 
  halign=left, valign=left}
  
\begin{tcolorbox}[colframe=gray!40!black, colback=gray!5!white, title=Prompts for Humor Interpretation Task]
{\small

\textbf{System prompt}: \textit{You are a humorous assistant that understands comics. You will be given comics and your task is to evaluate the comics.}
\\ \\
\textbf{Question 6}: \textit{Explain why this comic is funny or not funny in 3 sentences.}

}
\end{tcolorbox}

\newtcolorbox[auto counter, number within=section]{mutatorbox8}[2][]{colback=gray!10!white, colframe=black, coltitle=black, title=Mutator Prompts~\thetcbcounter: #2,#1, sharp corners, boxrule=0.5mm, left=10mm, right=10mm, fonttitle=\bfseries, 
  halign=left, valign=left}
  
\begin{tcolorbox}[colframe=gray!40!black, colback=gray!5!white, title=Prompts for Sequence Recognition Task]
{\small

\textbf{System prompt}: \textit{You are a humorous assistant that understands comics. You will be given comics and your task is to evaluate the comics.}
\\ \\
\textbf{Question 7}: \textit{In what order should the panels be read? Respond with the panel numbers only. Write the panel numbers followed by a comma. For example the answer ``3,4,2,1'' will mean that panels will be read in order of  panel 3, then panel 4, then panel 2 and finally panel 1.}
\\ \\
\textbf{Question 8}: \textit{For each panel, what are the text inside? Respond as \{panel\textunderscore number\}: \{text\textunderscore within\textunderscore panel\}.}

}
\label{apx:sequence_recognition_prompt}
\end{tcolorbox}

\section{Quantitative Error Analysis}
\label{apx:quantitative_error_analysis}
We have also conducted a structured, quantitative error analysis by (i) providing confusion matrices for humor style classification across models, and (ii) presenting F1 score breakdowns across five reasoning tasks, segmented by humor style and number of comic panels. These allow us to identify patterns in model reasoning failures, for instance, sarcasm and surreal humor consistently present challenges across models, and performance tends to degrade as panel count increases. This deeper diagnostic view supports targeted directions for future work, such as better context modeling and culturally-informed reasoning modules.

Table \ref{tab:confusion_matrix_humor_style} presents the confusion matrix for humor style classification task for each model. From the confusion matrix, we can see that models classified each of the comics to have multiple humor styles according to Table \ref{tab:humor_style_num} predicted larger amount of False Positive classification, which lead to a lower precision score and hence low F1 score. On the other hand, LLaVA-OV 7B which only predicted 1 humor style per comic has larger amount of False Negatives while most classifications on Sarcasm are actually False Positives.

\begin{table*}
    \centering
    \begin{tabular}{|c|c|c|c|c||c|c|c|c||c|c|c|c|}
        \hline
        Model & \multicolumn{4}{|c||}{GPT-4o} & \multicolumn{4}{|c||}{Gemini-1.5-Pro} & \multicolumn{4}{|c|}{Qwen2-VL-72B}\\ \hline
        Styles & TP & FP & TN & FN & TP & FP & TN & FN & TP & FP & TN & FN \\ \hline \hline
        Com. & 143 & 364 & 2196 & 97  & 101 & 144  & 2416 & 139 & 73 & 69 & 2491 & 167 \\
        Per. & 625 & 437 & 1715 & 23  & 531 & 305  & 1847 & 117 & 544 & 296 & 1856 & 104 \\
        Exa. & 383 & 864 & 1431 & 122 & 243 & 437  & 1858 & 262 & 263 & 515 & 1780 & 242  \\
        Pun  & 132 & 220 & 2355 & 93 & 106 & 243  & 2332 & 119 & 92 & 221 & 2354 & 133 \\
        Sar. & 152 & 571 & 1962 & 115 & 228 & 1578 & 955  & 39  & 67 & 314 & 2219 & 200 \\
        Sil. & 393 & 793 & 1344 & 270 & 434 & 1049 & 1088 & 229 & 177 & 342 & 1795 & 486  \\
        Sur. & 647 & 778 & 1114 & 261 & 716 & 978  & 914  & 192 & 116 & 88 & 1804 & 792 \\
        Dar. & 129 & 356 & 2271 & 44 & 101 & 207  & 2420 & 72  & 62 & 61 & 2566 & 111  \\
        N/A  & 1 & 1 & 2766 & 32 & 3   & 5    & 2762 & 30  & 6 & 12 & 2755 & 27 \\
        \hline
    \end{tabular}

    \vspace{0.5cm}

    \begin{tabular}{|c|c|c|c|c||c|c|c|c||c|c|c|c|}
        \hline
        Model & \multicolumn{4}{|c||}{Gemma3-27B} & \multicolumn{4}{|c||}{LLaVA-OV 7B} & \multicolumn{4}{|c|}{Qwen2-VL 7B}\\ \hline
        Styles & TP & FP & TN & FN & TP & FP & TN & FN & TP & FP & TN & FN \\ \hline \hline
        Com. & 107 & 186  & 2374 & 133 & 93 & 302 & 2258 & 147 & 112 & 414 & 2146 & 128 \\
        Per. & 497 & 293  & 1859 & 151 & 86 & 199 & 2096 & 419 & 234 & 624 & 1671 & 271 \\
        Exa. & 291 & 580  & 1715 & 214 & 46 & 13 & 2139 & 602 & 568 & 347 & 1805 & 80 \\
        Pun  & 75  & 114  & 2461 & 150 & 17 & 63 & 2512 & 208 & 61 & 323 & 2252 & 164 \\
        Sar. & 184 & 1120 & 1413 & 83  & 207 & 1765 & 768 & 60 & 27 & 185 & 2348 & 240 \\
        Sil. & 607 & 1813 & 324  & 56  & 3 & 3 & 2134 & 660 & 36 & 79 & 2058 & 627 \\
        Sur. & 784 & 1278 & 614  & 124 & 0 & 1 & 1891 & 908 & 1 & 0 & 1892 & 907 \\
        Dar. & 143 & 605  & 2022 & 30  & 0 & 0 & 2627 & 173 & 41 & 53 & 2574 & 132 \\
        N/A  & 3   & 2    & 2765 & 30  & 0 & 2 & 2765 & 33 & 1 & 4 & 2763 & 32 \\
        \hline
    \end{tabular}
    \caption{Confusion matrix of the model prediction on the Humor Classification task broken down by humor style. TP: True Positive, FP: False Positive, TN: True Negative, FN: False Negative.}
    \label{tab:confusion_matrix_humor_style}
\end{table*}

Table \ref{tab:f1_humor_style} shows the F1 scores for each task broken down by humor styles for each model. In Task 3, panel contribution task, all tested models have better performance on comics with Personification as one of the humor styles as personification is more straightforward to identify than other humor styles. On the other hand, Exaggeration performs worse than most of the other humor styles. For Task 4, modality contribution task, most of the models (except Gemma3-27B and Qwen2-VL-7B) can identify the modality contributing to sarcastic comic better. While interestingly most of them failed to correctly identify the modality contribution for those comics with personification as humor style. (RY: I didn't include for Task 5, since it seems to be repeating Table \ref{tab:humor_classification_results}, but that table is Recall instead of F1.)

\begin{table*}
    \small
    \centering
    \begin{tabular}{|c|c|c|c|c|c||c|c|c|c|c|}
        \hline
        Model & \multicolumn{5}{|c||}{GPT-4o} & \multicolumn{5}{|c|}{Gemini-1.5-Pro} \\ \hline
        Styles & Task 1 & Task 2 & Task 3 & Task 4 & Task 5 & Task 1 & Task 2 & Task 3 & Task 4 & Task 5 \\ \hline \hline
        Com. & 1.000 & 0.911 & 0.760 & 0.690 & 0.594 & 0.998 & 0.878 & 0.683 & 0.649 & 0.477 \\
        Per. & 1.000 & 0.928 & 0.835 & 0.512 & 0.763 & 1.000 & 0.879 & 0.772 & 0.525 & 0.690 \\
        Exa. & 1.000 & 0.890 & 0.681 & 0.621 & 0.660 & 0.999 & 0.844 & 0.661 & 0.611 & 0.522 \\
        Pun & 1.000 & 0.935 & 0.773 & 0.669 & 0.609 & 1.000 & 0.904 & 0.778 & 0.661 & 0.544 \\
        Sar. & 1.000 & 0.910 & 0.777 & 0.764 & 0.586 & 0.998 & 0.883 & 0.741 & 0.749 & 0.708 \\
        Sil. & 1.000 & 0.921 & 0.777 & 0.624 & 0.605 & 0.999 & 0.870 & 0.713 & 0.583 & 0.630 \\
        Sur. & 0.999 & 0.901 & 0.782 & 0.641 & 0.664 & 0.999 & 0.870 & 0.728 & 0.637 & 0.692 \\
        Dar. & 0.997 & 0.888 & 0.764 & 0.576 & 0.688 & 0.997 & 0.829 & 0.765 & 0.556 & 0.610 \\
        N/A & 0.059 & 0.985 & 0.059 & 0.059 & 0.059 & 0.167 & 0.935 & 0.167 & 0.167 & 0.167 \\
        \hline
    \end{tabular}

    \vspace{0.5cm}

    \begin{tabular}{|c|c|c|c|c|c||c|c|c|c|c|}
        \hline
        Model & \multicolumn{5}{|c||}{Qwen2-VL-72B} & \multicolumn{5}{|c|}{Gemma3-27B} \\ \hline
        Styles & Task 1 & Task 2 & Task 3 & Task 4 & Task 5 & Task 1 & Task 2 & Task 3 & Task 4 & Task 5 \\ \hline \hline
        Com. & 1.000 & 0.822 & 0.528 & 0.647 & 0.415 & 1.000 & 0.836 & 0.547 & 0.229 & 0.504 \\
        Per. & 1.000 & 0.784 & 0.531 & 0.456 & 0.678 & 1.000 & 0.812 & 0.597 & 0.311 & 0.665 \\
        Exa. & 0.998 & 0.735 & 0.462 & 0.611 & 0.513 & 1.000 & 0.787 & 0.474 & 0.171 & 0.563 \\
        Pun & 0.998 & 0.792 & 0.434 & 0.617 & 0.472 & 1.000 & 0.824 & 0.536 & 0.171 & 0.442 \\
        Sar. & 0.998 & 0.811 & 0.474 & 0.651 & 0.353 & 1.000 & 0.832 & 0.531 & 0.118 & 0.622 \\
        Sil. & 0.998 & 0.765 & 0.508 & 0.556 & 0.381 & 0.999 & 0.808 & 0.560 & 0.233 & 0.737 \\
        Sur. & 0.997 & 0.751 & 0.443 & 0.582 & 0.253 & 0.999 & 0.790 & 0.502 & 0.186 & 0.716 \\
        Dar. & 0.988 & 0.747 & 0.468 & 0.501 & 0.415 & 0.994 & 0.805 & 0.535 & 0.198 & 0.692 \\
        N/A & 0.308 & 0.862 & 0.308 & 0.308 & 0.308 & 0.167 & 0.900 & 0.167 & 0.167 & 0.167 \\
        \hline
    \end{tabular}

    \vspace{0.5cm}

    \begin{tabular}{|c|c|c|c|c|c||c|c|c|c|c|}
        \hline
        Model & \multicolumn{5}{|c||}{LLaVA-OV-7B} & \multicolumn{5}{|c|}{Qwen2-VL-7B} \\ \hline
        Styles & Task 1 & Task 2 & Task 3 & Task 4 & Task 5 & Task 1 & Task 2 & Task 3 & Task 4 & Task 5 \\ \hline \hline
        Com. & 1.000 & 0.761 & 0.459 & 0.605 & 0.394 & 1.000 & 0.042 & 0.575 & 0.240 & 0.488 \\
        Per. & 1.000 & 0.740 & 0.552 & 0.485 & 0.111 & 1.000 & 0.029 & 0.570 & 0.281 & 0.663 \\
        Exa. & 0.999 & 0.719 & 0.479 & 0.589 & 0.200 & 0.999 & 0.029 & 0.474 & 0.207 & 0.439 \\
        Pun & 1.000 & 0.805 & 0.463 & 0.563 & 0.098 & 1.000 & 0.017 & 0.450 & 0.232 & 0.340 \\
        Sar. & 1.000 & 0.791 & 0.498 & 0.656 & 0.582 & 1.000 & 0.025 & 0.509 & 0.075 & 0.208 \\
        Sil. & 0.999 & 0.754 & 0.512 & 0.583 & 0.030 & 0.998 & 0.048 & 0.547 & 0.257 & 0.140 \\
        Sur. & 0.999 & 0.757 & 0.480 & 0.565 & 0.028 & 0.999 & 0.025 & 0.455 & 0.200 & 0.082 \\
        Dar. & 1.000 & 0.761 & 0.523 & 0.453 & 0.008 & 0.994 & 0.026 & 0.431 & 0.336 & 0.253 \\
        N/A & 0.000 & 0.862 & 0.000 & 0.000 & 0.000 & 0.059 & 0.000 & 0.059 & 0.059 & 0.059 \\
        \hline
    \end{tabular}
    \caption{F1 score broken down by humor style for each task. Task 1: Humor Presence, Task 2: Sound Effect presence, Task 3: Panel Contribution, Task 4: Modality Contribution, Task 5: Humor Style Classification.}
    \label{tab:f1_humor_style}
\end{table*}

Table \ref{tab:f1_panel_num} shows the F1 scores for each task broken down by number of panels for each model. For Task 3 to Task 5, we can see that when the number of panels increases, the model performance degrades, showing the model's capability in understanding long-form humor remains challenging.

\begin{table*}
    \small
    \centering
    \begin{tabular}{|c|c|c|c|c|c||c|c|c|c|c|}
        \hline
        Model & \multicolumn{5}{|c||}{GPT-4o} & \multicolumn{5}{|c|}{Gemini-1.5-Pro} \\ \hline
        \# Panels & Task 1 & Task 2 & Task 3 & Task 4 & Task 5 & Task 1 & Task 2 & Task 3 & Task 4 & Task 5 \\ \hline \hline
        1  & 0.959 & 0.976 & 0.870 & 0.718 & 0.495 & 0.961 & 0.945 & 0.785 & 0.672 & 0.456 \\
        2  & 0.978 & 0.933 & 0.887 & 0.542 & 0.594 & 0.978 & 0.925 & 0.807 & 0.540 & 0.564 \\
        3  & 0.993 & 0.919 & 0.827 & 0.584 & 0.479 & 0.993 & 0.881 & 0.810 & 0.562 & 0.458 \\
        4  & 0.987 & 0.913 & 0.757 & 0.641 & 0.524 & 0.987 & 0.871 & 0.721 & 0.636 & 0.504 \\
        5  & 0.967 & 0.829 & 0.538 & 0.567 & 0.528 & 0.967 & 0.788 & 0.556 & 0.575 & 0.511 \\
        6  & 1.000 & 0.886 & 0.728 & 0.598 & 0.493 & 1.000 & 0.817 & 0.665 & 0.572 & 0.538 \\
        7  & 1.000 & 0.928 & 0.658 & 0.598 & 0.470 & 1.000 & 0.818 & 0.681 & 0.616 & 0.494 \\
        8  & 0.972 & 0.792 & 0.595 & 0.486 & 0.477 & 0.972 & 0.766 & 0.603 & 0.581 & 0.481 \\
        9  & 0.926 & 1.000 & 0.567 & 0.486 & 0.470 & 0.926 & 0.952 & 0.242 & 0.617 & 0.451 \\
        10 & 1.000 & 0.855 & 0.355 & 0.609 & 0.576 & 1.000 & 0.757 & 0.414 & 0.639 & 0.427 \\
        11 & 1.000 & 0.904 & 0.133 & 0.630 & 0.399 & 1.000 & 0.836 & 0.207 & 0.778 & 0.487 \\
        12 & 1.000 & 1.000 & 0.667 & 0.556 & 0.875 & 1.000 & 0.667 & 0.667 & 0.556 & 0.625 \\
        13 & 1.000 & 0.333 & 0.000 & 0.333 & 0.667 & 1.000 & 0.333 & 0.500 & 0.333 & 0.667 \\
        14 & 1.000 & 1.000 & 0.000 & 0.000 & 1.000 & 1.000 & 1.000 & 0.000 & 0.000 & 1.000 \\
        17 & 1.000 & 1.000 & 0.000 & 0.000 & 1.000 & 1.000 & 0.000 & 1.000 & 0.000 & 1.000 \\
        18 & 1.000 & 1.000 & 1.000 & 1.000 & 0.000 & 1.000 & 1.000 & 0.000 & 1.000 & 0.000 \\
        \hline
    \end{tabular}

    \vspace{0.5cm}

    \begin{tabular}{|c|c|c|c|c|c||c|c|c|c|c|}
        \hline
        Model & \multicolumn{5}{|c||}{Qwen2-VL-72B} & \multicolumn{5}{|c|}{Gemma3-27B} \\ \hline
        \# Panels & Task 1 & Task 2 & Task 3 & Task 4 & Task 5 & Task 1 & Task 2 & Task 3 & Task 4 & Task 5 \\ \hline \hline
        1  & 0.969 & 0.900 & 0.844 & 0.675 & 0.372 & 0.966 & 0.907 & 0.876 & 0.345 & 0.461 \\
        2  & 0.974 & 0.820 & 0.600 & 0.549 & 0.450 & 0.978 & 0.840 & 0.529 & 0.337 & 0.590 \\
        3  & 0.992 & 0.735 & 0.407 & 0.473 & 0.347 & 0.993 & 0.786 & 0.533 & 0.185 & 0.433 \\
        4  & 0.987 & 0.779 & 0.364 & 0.615 & 0.373 & 0.987 & 0.823 & 0.416 & 0.160 & 0.492 \\
        5  & 0.967 & 0.631 & 0.386 & 0.586 & 0.396 & 0.967 & 0.652 & 0.306 & 0.197 & 0.478 \\
        6  & 0.994 & 0.705 & 0.302 & 0.425 & 0.387 & 1.000 & 0.735 & 0.375 & 0.172 & 0.476 \\
        7  & 1.000 & 0.687 & 0.474 & 0.579 & 0.340 & 1.000 & 0.748 & 0.410 & 0.160 & 0.461 \\
        8  & 0.962 & 0.708 & 0.466 & 0.450 & 0.273 & 0.972 & 0.694 & 0.332 & 0.172 & 0.460 \\
        9  & 0.926 & 0.863 & 0.383 & 0.664 & 0.356 & 0.926 & 0.952 & 0.197 & 0.181 & 0.465 \\
        10 & 1.000 & 0.563 & 0.268 & 0.641 & 0.455 & 1.000 & 0.633 & 0.205 & 0.236 & 0.487 \\
        11 & 1.000 & 0.728 & 0.133 & 0.519 & 0.462 & 1.000 & 0.904 & 0.211 & 0.274 & 0.398 \\
        12 & 1.000 & 0.667 & 0.444 & 0.556 & 0.650 & 1.000 & 1.000 & 0.000 & 0.167 & 0.400 \\
        13 & 1.000 & 0.333 & 0.500 & 0.333 & 0.250 & 1.000 & 0.333 & 0.000 & 0.333 & 0.500 \\
        14 & 1.000 & 1.000 & 1.000 & 1.000 & 1.000 & 1.000 & 1.000 & 1.000 & 0.000 & 1.000 \\
        17 & 1.000 & 0.000 & 0.000 & 0.000 & 1.000 & 1.000 & 0.000 & 0.000 & 0.000 & 1.000 \\
        18 & 1.000 & 1.000 & 0.000 & 1.000 & 0.000 & 1.000 & 1.000 & 0.000 & 1.000 & 0.000 \\
        \hline
    \end{tabular}

    \vspace{0.5cm}

    \begin{tabular}{|c|c|c|c|c|c||c|c|c|c|c|}
        \hline
        Model & \multicolumn{5}{|c||}{LLaVA-OV-7B} & \multicolumn{5}{|c|}{Qwen2-VL-7B} \\ \hline
        \# Panels & Task 1 & Task 2 & Task 3 & Task 4 & Task 5 & Task 1 & Task 2 & Task 3 & Task 4 & Task 5 \\ \hline \hline
        1  & 0.952 & 0.838 & 0.763 & 0.587 & 0.123 & 0.952 & 0.076 & 0.939 & 0.286 & 0.272 \\
        2  & 0.978 & 0.768 & 0.442 & 0.535 & 0.164 & 0.978 & 0.000 & 0.801 & 0.341 & 0.383 \\
        3  & 0.993 & 0.745 & 0.421 & 0.514 & 0.087 & 0.996 & 0.025 & 0.391 & 0.186 & 0.241 \\
        4  & 0.988 & 0.753 & 0.441 & 0.586 & 0.095 & 0.987 & 0.014 & 0.347 & 0.167 & 0.254 \\
        5  & 0.967 & 0.670 & 0.458 & 0.544 & 0.071 & 0.967 & 0.011 & 0.285 & 0.247 & 0.162 \\
        6  & 1.000 & 0.732 & 0.365 & 0.489 & 0.026 & 1.000 & 0.003 & 0.124 & 0.264 & 0.162 \\
        7  & 1.000 & 0.782 & 0.359 & 0.515 & 0.051 & 1.000 & 0.001 & 0.303 & 0.198 & 0.268 \\
        8  & 0.972 & 0.708 & 0.448 & 0.489 & 0.019 & 0.972 & 0.079 & 0.221 & 0.242 & 0.115 \\
        9  & 0.926 & 0.907 & 0.193 & 0.586 & 0.165 & 0.926 & 0.080 & 0.115 & 0.172 & 0.244 \\
        10 & 1.000 & 0.637 & 0.139 & 0.596 & 0.114 & 1.000 & 0.074 & 0.135 & 0.330 & 0.323 \\
        11 & 1.000 & 0.904 & 0.000 & 0.474 & 0.000 & 1.000 & 0.000 & 0.000 & 0.316 & 0.460 \\
        12 & 1.000 & 1.000 & 0.444 & 0.222 & 0.000 & 1.000 & 0.000 & 0.444 & 0.167 & 0.400 \\
        13 & 1.000 & 0.333 & 0.500 & 0.333 & 0.167 & 1.000 & 0.500 & 0.000 & 0.333 & 0.417 \\
        14 & 1.000 & 1.000 & 0.000 & 0.000 & 0.000 & 1.000 & 0.000 & 1.000 & 0.000 & 0.000 \\
        17 & 1.000 & 0.000 & 0.000 & 0.000 & 0.000 & 1.000 & 0.000 & 0.000 & 1.000 & 1.000 \\
        18 & 1.000 & 1.000 & 0.000 & 1.000 & 0.000 & 1.000 & 0.000 & 0.000 & 1.000 & 0.000 \\
        \hline
    \end{tabular}
    \caption{F1 score broken down by number of panels for each task. Task 1: Humor Presence, Task 2: Sound Effect presence, Task 3: Panel Contribution, Task 4: Modality Contribution, Task 5: Humor Style Classification. Note that there are no comics with 15 and 16 panels.}
    \label{tab:f1_panel_num}
\end{table*}

\section{Case Studies}
\label{apx:case_study}
In all case studies,  the XKCD comics are used for non-commercial purposes under the Creative Commons Attribution-NonCommercial 2.5 License.

Table~\ref{tab:case_studies} presents a case where models struggled with humor identification and classification but performed well in humor interpretation. While LLaVA-OV achieved the highest scores in both tasks, its explanation was widely disagreed upon by annotators, suggesting that high classification accuracy does not necessarily correlate with meaningful humor interpretation. In contrast, other models with lower classification scores provided interpretations that made more sense to human evaluators, highlighting the complexity of aligning AI-generated humor reasoning with human perception.

A common failure in sequence recognition is that models default to conventional reading orders, incorrectly associating speech bubbles with different panels. Table~\ref{tab:case_studies_2} illustrates how models fail to fully grasp context, even when they correctly recognize text. Both Gemini-1.5-Pro and LLaVA-OV produced hallucinated responses, introducing details absent from the comic, such as body language and facial expressions. Meanwhile, GPT-4o and Qwen2-VL failed to capture the punchline, missing the subtle wordplay on "time travel" and the gradual realization conveyed in the comic. Even when GPT-4o correctly identified the panel sequence and text, it struggled to connect them to the underlying humor, reinforcing that sequential reasoning alone is insufficient for humor comprehension.

Table~\ref{tab:case_studies_3} presents an intriguing contrast. GPT-4o misidentified the panel sequence but correctly mapped text to panels, producing an interpretation that annotators found reasonable. In contrast, Gemini-1.5-Pro correctly aligned the panels and text but failed to provide a humor explanation that resonated with human evaluators. This suggests that perfect sequence recognition may not be essential for humor comprehension, raising questions about the relative importance of narrative alignment versus inferential reasoning. Further studies are needed to explore this phenomenon and determine whether narrative structure alone is sufficient for AI to understand humor, or if deeper contextual reasoning is required.

\onecolumn
\begin{center}
\small
\begin{longtable}{|l|p{1cm}|p{1cm}|l|p{1.2cm}|p{1.4cm}|p{5cm}|}
\caption{Case study 1 on an XKCD comic, access at \url{https://xkcd.com/234/}. Incorrect classifications are marked in \red{red}.}
\label{tab:case_studies} \\

\hline
\multicolumn{7}{|c|}{\includegraphics[width=0.8\linewidth]{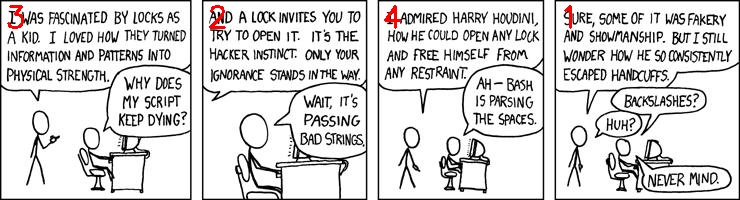}} \\ \hline
\textbf{Model} & \textbf{Humor} & \textbf{Sound} & \textbf{Punchline Panel} & \textbf{Modality} & \textbf{Humor Type} & \textbf{Interpretation} \\ \hline 
\noalign{\vskip 3pt}
\hline
Groundtruth     & Yes & Yes & 1 & Text & Comparison & - \\ \hline
\endhead

\hline \multicolumn{7}{r}{{\small\it Continued on next page}} \\ \hline
\endfoot

\hline
\endlastfoot

GPT-4o          & Yes & \red{NA} & \red{5} & Text & \red{Pun} & This comic is funny because it cleverly combines the concept of lock-picking with programming, using technical jargon as a metaphor for escaping restraints. The punchline involves a play on words with "backslashes," which are used in coding, humorously suggesting they could help escape handcuffs. The humor lies in the absurdity of applying programming logic to a physical escape scenario. \newline
Score: [6,6] \\ \hline
Gemini-1.5-Pro  & Yes & \red{NA} & \red{2} & Text & \red{Pun, Silliness} & This comic is mildly humorous due to the juxtaposition of romanticized lock-picking with the mundane reality of debugging code.  The stick figure's script crashing is a relatable experience for programmers.  The punchline about backslashes adds a layer of humor specific to coding, further enhancing the relatability for a niche audience. \newline
Score: [4,5] \\ \hline
Qwen2-VL 72B  & Yes & \red{NA} & \red{4} & Text & \red{Pun} & This comic is funny because it cleverly juxtaposes the fascination with locks and the hacker instinct with the mundane frustration of debugging code. The humor lies in the contrast between the grandiosity of admiring Houdini's lock-picking skills and the petty annoyance of dealing with spaces in a Bash script. The punchline, where the character dismissively says \"Backslashes? Huh? Never mind,\" adds to the humor by highlighting the sudden realization that the problem was trivial and easily solved. \newline
Score: [2,3] \\ \hline
Gemma3 27B  & Yes & \red{NA} & \red{2} & \red{Both} & Comparison\red{, Exaggeration, Silliness} & Ah, a classic XKCD! This comic is funny because it perfectly captures the frustrating experience of debugging code – starting with grand philosophical thoughts about locks and hacking, then spiraling down into the mundane reality of whitespace errors. The humor lies in the relatable contrast between the lofty aspirations of a programmer and the petty, infuriating bugs that plague their work, culminating in the realization that the problem was just a backslash. It's a joke that only someone who's stared at a computer screen for too long will *really* get, making it a delightfully nerdy chuckle. \newline
Score: [3,2] \\ \hline
LLaVA-OV 7B     & Yes & Yes & \red{5} & Text & Comparison & The comic is funny because it plays on the idea of a person being fascinated by locks and trying to open them, only to realize that they are not as strong as they thought. The humor comes from the person's realization and the absurdity of the situation. Additionally, the comic includes a twist at the end where the person discovers that the lock was fake, adding to the comedic effect.\newline
Score: [1,1] \\ \hline
Qwen2-VL 7B     & Yes & \red{NA} & \red{2} & \red{Both} & \red{Pun} & This comic is funny because it uses humor to highlight the challenges and frustrations of learning to code. The conversation between the two characters is relatable to anyone who has struggled with coding, and the punchline is unexpected and amusing. The comic also uses visual humor, with the character's expressions and body language adding to the humor. \newline
Score: [2,4]  \\ \hline
\end{longtable}
\end{center}

\onecolumn
\begin{center}
\small
\begin{longtable}{|l|p{1.5cm}|p{5cm}|p{5cm}|}
\caption{Case study 2 on an XKCD comic, access at \url{https://xkcd.com/209/}. Incorrect recognitions are marked in \red{red}.}
\label{tab:case_studies_2} \\

\hline
\multicolumn{4}{|c|}{\includegraphics[width=0.8\linewidth]{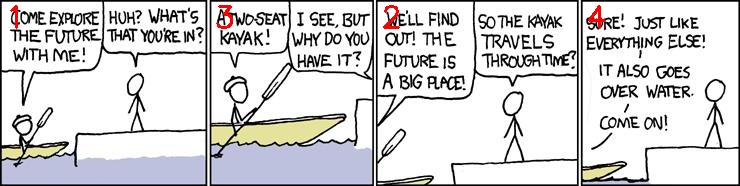}} \\ \hline
\textbf{Model} & \textbf{Sequence Identification} & \textbf{Text Identification} & \textbf{Interpretation} \\ \hline
\noalign{\vskip 3pt}
\hline
Groundtruth     & 1,3,2,4 & 1: COME EXPLORE THE FUTURE WITH ME! HUH? WHAT'S THAT YOU'RE IN? 3: TWO-SEAT KAYAK! I SEE, BUT WHY DO YOU HAVE IT? 2: WE'LL FIND OUT! THE FUTURE IS A BIG PLACE! SO THE KAYAK TRAVELS THROUGH TIME? 4: SURE! JUST LIKE EVERYTHING ELSE! IT ALSO GOES OVER WATER. COME ON! & - \\ \hline
\endhead

\hline \multicolumn{4}{r}{{\small\it Continued on next page}} \\ \hline
\endfoot

\hline
\endlastfoot

GPT-4o          & 1,3,2,4 & 1: COME EXPLORE THE FUTURE WITH ME! \red{2:} HUH? WHAT'S THAT YOU'RE IN? 3: TWO-SEAT KAYAK! \red{4:} I SEE, BUT WHY DO YOU HAVE IT? \red{5:} WE'LL FIND OUT! THE FUTURE IS A BIG PLACE! \red{6:} SO THE KAYAK TRAVELS THROUGH TIME? \red{7:} SURE! JUST LIKE EVERYTHING ELSE! IT ALSO GOES OVER WATER. COME ON! & This comic is funny because it plays on the literal interpretation of exploring the future with a mundane object, a two-seat kayak. The humor lies in the absurdity of using a kayak, which travels over water, as a metaphor for time travel. The punchline cleverly highlights the character's enthusiasm and the unexpected twist of exploring the future in such an ordinary way. \newline
Score: [3,1] \\ \hline
Gemini-1.5-Pro  & \red{4,2,3,1} & \red{1:} COME EXPLORE THE FUTURE WITH ME! HUH? WHAT'S THAT YOU'RE IN? \red{2:} A TWO-SEAT KAYAK! I SEE, BUT WHY DO YOU HAVE IT? \red{3:} WE'LL FIND OUT! THE FUTURE IS A BIG PLACE! SO THE KAYAK TRAVELS THROUGH TIME? \red{4:} SURE! JUST LIKE EVERYTHING ELSE! IT ALSO GOES OVER WATER. COME ON! & This comic is funny because of the absurd premise of using a kayak to travel through time. The straight-faced delivery of the first character, despite the obvious lack of logic, adds to the humor. Finally, the second character's acceptance of this flawed logic makes the situation even more ridiculous and funny. \newline
Score: [1,1] \\ \hline
Qwen2-VL 72B   & \red{1,2,3,4} & 1: COME EXPLORE THE FUTURE WITH ME! \red{2:} HUH? WHAT'S THAT YOU'RE IN? 3: A TWO-SEAT KAYAK! \red{4:} I SEE, BUT WHY DO YOU HAVE IT? \red{5:} WE'LL FIND OUT! THE FUTURE IS A BIG PLACE! \red{6:} SO THE KAYAK TRAVELS THROUGH TIME? \red{7:} SURE! JUST LIKE EVERYTHING ELSE! IT ALSO GOES OVER WATER. COME ON! & This comic is funny because it plays on the idea of exploring the future, which is typically associated with time travel. The punchline reveals that the "future" the character is referring to is actually just a kayak trip, which is a clever and unexpected twist. The humor comes from the contrast between the grandiose idea of exploring the future and the mundane reality of a kayak ride. \newline
Score: [3,6] \\ \hline
Gemma3 27B   & \red{1,2,3,4} & 1: COME EXPLORE \red{HUH? WHAT'S} THE FUTURE \red{THAT YOU'RE IN} WITH ME! \red{2: WOAHSEAT,} I SEE, BUT WHY DO YOU HAVE IT? \red{3}: WE'LL FIND OUT! THE FUTURE IS A BIG PLACE! 4: \red{SO THE KAYAK TRAVELS THROUGH TIME!} SURE. JUST LIKE EVERYTHING ELSE! IT ALSO GOES OVER WATER. COME ON! & Ah, a classic xkcd! This comic is funny because it takes a wildly complex concept - time travel - and applies the most mundane, literal solution: a kayak. The humor comes from the deadpan delivery and the complete disregard for any actual scientific principles of time travel, culminating in the realization that, of course, a kayak also floats on water. It's the kind of joke that makes you chuckle and then question your own logic. \newline
Score: [3,3] \\ \hline
LLaVA-OV 7B     & \red{1,2,3,4} & \red{1:} Come explore the future with me! \red{2:} Uh? What's that you're in? 3: I see, but why do you have it? \red{4:} So the kayak travels through time? \red{It also goes over water. Come on!} & The comic is funny because it plays on the idea of a person trying to explain the concept of time travel to someone who is not familiar with it. The humor comes from the absurdity of the situation and the person's frustration in trying to convey a complex idea in simple terms. Additionally, the comic uses visual humor by showing the person's body language and facial expressions as they struggle to explain the concept.\newline
Score: [1,1] \\ \hline
Qwen2-VL 7B     & \red{1,2,3,4} & \red{1:} COME EXPLORE THE FUTURE WITH ME! \red{2:} HUH? WHAT'S THAT YOU'RE IN? 3: A TWO-SEAT KAYAK! \red{4:} I SEE, BUT WHY DO YOU HAVE IT? \red{5:} WE'LL FIND OUT! THE FUTURE IS A BIG PLACE! \red{6:} SO THE KAYAK TRAVELS THROUGH TIME? \red{7:} SURE! JUST LIKE EVERYTHING ELSE! IT ALSO GOES OVER WATER. COME ON! & This comic is funny because it plays on the idea of time travel and the absurdity of a two-seat kayak. The dialogue between the characters is also humorous, with the first character's offer to explore the future and the second character's confusion about the kayak's purpose. The final panel, with the second character's response, adds to the humor by emphasizing the absurdity of the situation. \newline
Score: [3,3]  \\ \hline
\end{longtable}
\end{center}

\onecolumn
\begin{center}
\small
\begin{longtable}{|l|p{1.5cm}|p{5cm}|p{5cm}|}
\caption{Case study 3 on an XKCD comic, access at \url{https://xkcd.com/473/}. Incorrect recognitions are marked in \red{red}.}
\label{tab:case_studies_3} \\

\hline
\multicolumn{4}{|c|}{\includegraphics[width=0.8\linewidth]{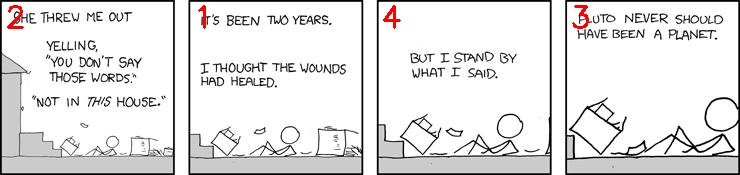}} \\ \hline
\textbf{Model} & \textbf{Sequence Identification} & \textbf{Text Identification} & \textbf{Interpretation} \\ \hline
\noalign{\vskip 3pt}
\hline
Groundtruth     & 2,1,4,3 & 2: SHE THREW ME OUT YELLING, "YOU DON'T SAY THOSE WORDS. "NOT IN THIS HOUSE." 1: IT'S BEEN TWO YEARS. I THOUGHT THE WOUNDS HAD HEALED. 4: BUT I STAND BY WHAT I SAID. 3: PLUTO NEVER SHOULD HAVE BEEN A PLANET. & - \\ \hline
\endhead

\hline \multicolumn{4}{r}{{\small\it Continued on next page}} \\ \hline
\endfoot

\hline
\endlastfoot

GPT-4o          & \red{1,2,3,4} & 1: It's been two years. I thought the wounds had healed. 2: She threw me out yelling, "You don't say those words." "Not in this house." 3: Pluto never should have been a planet. 4: But I stand by what I said. & This comic is funny because it humorously exaggerates the emotional fallout from a scientific debate about Pluto's planetary status. The dramatic setup, akin to a personal or familial conflict, contrasts with the seemingly trivial nature of the argument over Pluto. The punchline reveals the absurdity of being thrown out of a house over an astronomical opinion, highlighting the humor in taking such a stance too seriously. \newline
Score: [6,6] \\ \hline
Gemini-1.5-Pro  & 2,1,4,3 & 1: IT'S BEEN TWO YEARS. I THOUGHT THE WOUNDS HAD HEALED. 2: \red{HE} THREW ME OUT YELLING, "YOU DON'T SAY THOSE WORDS. "NOT IN THIS HOUSE." 3: PLUTO NEVER SHOULD HAVE BEEN A PLANET. 4: BUT I STAND BY WHAT I SAID. & This comic is funny because it highlights the stubbornness of people and their unwillingness to admit they're wrong, even after a long time. The visual gag of the character remaining in the same position outside the house for two years adds to the humor. The absurdity of the situation, particularly the final panel's non-sequitur about Pluto, elevates the comic further. \newline
Score: [1,2] \\ \hline
Qwen2-VL 72B  & \red{1,2,3,4} & 1: IT'S BEEN TWO YEARS. 2: SHE THREW ME OUT, YELLING, "YOU DON'T SAY THOSE WORDS." "NOT IN THIS HOUSE." \red{3}: BUT I STAND BY WHAT I SAID. \red{4}: PLUTO NEVER SHOULD HAVE BEEN A PLANET. & This comic is funny because it plays on the absurdity of the situation. The first panel sets up a serious and dramatic tone, suggesting a deep familial conflict. However, the punchline in the last panel reveals that the argument was over the demotion of Pluto from planet status, which is a trivial and unexpected topic for such a heated argument. The contrast between the serious setup and the silly resolution creates a humorous effect. \newline
Score: [7,7] \\ \hline
Gemma3 27B  & \red{1,2,3,4} & 1: IT'S BEEN TWO YEARS. I THOUGHT THE WOUNDS HAD HEALED. 2: SHE THREW ME OUT YELLING, "YOU DON'T SAY THOSE WORDS." "NOT IN THIS HOUSE." 3: \red{AUTO} NEVER SHOULD HAVE BEEN A PLANET. 4: BUT I STAND BY WHAT I SAID. & Alright, let's dissect this xkcd gem! This comic is funny because it takes a dramatic, relationship-ending argument (\"You don't say those words! Not in this house!\") and then *completely* undercuts it with a bizarre, non-sequitur declaration about Pluto's planetary status. The escalation from domestic dispute to astronomical opinion is so unexpected and absurd that it lands as a delightfully nerdy punchline, and the visual of the aftermath just adds to the chaotic energy. It's a perfect example of xkcd's signature blend of relatable situations and delightfully strange tangents. \newline
Score: [7,7] \\ \hline
LLaVA-OV 7B     & 2,1,4,3 & 2: \red{HE} THREW ME OUT YELLING, "YOU DON'T SAY THOSE WORDS!" NOT IN THIS HOUSE. \red{3: BUT I STAND BY WHAT I SAID.} 4: BUT I STAND BY WHAT I SAID. & The comic is funny because it plays on the idea of a person who has been hurt by someone else's words, but then realizes that they have been using those same words to hurt others. The humor comes from the irony of the situation and the unexpected twist at the end. However, some people may not find it funny if they do not understand the context or if they have experienced similar situations in their own lives. \newline
Score: [1,1] \\ \hline
Qwen2-VL 7B     & 2,1,4,3 & 1: It's been two years. 2: She threw me out yelling, "You don't say those words." "Not in this house." 3: Pluto never should have been a planet. 4: But I stand by what I said. & This comic is funny because it uses a series of panels to depict a humorous and relatable situation. The first panel sets up the premise, and the subsequent panels build on it with increasingly absurd and exaggerated actions. The final panel, which shows the character being thrown out of a house, is the punchline that ties everything together in a humorous way. The use of simple, stick-figure drawings and minimal text also adds to the humor by making the situation seem more absurd and exaggerated. \newline
Score: [4,4] \\ \hline
\end{longtable}
\end{center}

%%%%%%%%%%%%%%%%%%%%%%%%%%%%%%%%%% Human Preference Appendix %%%%%%%%%%%%%%%%%%%%%%%%%%%%%%%%%% Formatting received helped from gpt 

% Compact boxes for explanations
\newcommand{\HumanBox}[1]{%
  \noindent\fbox{\parbox{\linewidth}{\textbf{Human interpretation:}~#1}}%
}
\newcommand{\ModelBoxA}[1]{%
  \noindent\fbox{\parbox{\linewidth}{\textbf{GPT4o Interpretation:}~#1}}%
}
\newcommand{\ModelBoxB}[1]{%
  \noindent\fbox{\parbox{\linewidth}{\textbf{Gpt4o interpretation:}~#1}}%
}

% % Helper to render a two-column row under an image
% \newcommand{\HumanBox}[1]{%
%   \noindent\fbox{\parbox{\linewidth}{\textbf{Human interpretation:}~#1}}%
% }
% \newcommand{\ModelBoxA}[1]{%
%   \noindent\fbox{\parbox{\linewidth}{\textbf{GPT4o Interpretation:}~#1}}%
% }
% \newcommand{\ModelBoxB}[1]{%
%   \noindent\fbox{\parbox{\linewidth}{\textbf{Gpt4o interpretation:}~#1}}%
% }

% Column type with vertical centering inside cells (improves alignment)
\newcolumntype{L}[1]{>{\raggedright\arraybackslash}m{#1}}

% Two-column row under an image with vertically centered cells
\newcommand{\AlignedExplainRowA}[2]{%
  \noindent\begin{tabular}{@{}L{0.49\linewidth}@{\hspace{0.02\linewidth}}L{0.49\linewidth}@{}}
    \small \HumanBox{#1} & \small \ModelBoxA{#2}
  \end{tabular}%
}
\newcommand{\AlignedExplainRowB}[2]{%
  \noindent\begin{tabular}{@{}L{0.49\linewidth}@{\hspace{0.02\linewidth}}L{0.49\linewidth}@{}}
    \small \HumanBox{#1} & \small \ModelBoxB{#2}
  \end{tabular}%
}

\section{Human Preference Examples}
\label{apx:human_explanation_examples}

% We further assessed the difficulty of humor interpretation for LMMs through a preference study. In this study, we selected 70 comics: 10 comics randomly chosen per source, and a human was tasked to write their interpretation of the humor in these comics. After which, two participants were tasked to choose the best interpretation of the humor---one human and six generated---for the comics. Disagreements are resolved through a third annotator from the chosen explanations.

Below are samples of human-written interpretations chosen over GPT4o generated interpretations. In these samples, we highlight spans where a human incorporated multimodal reasoning that contributed to the humor; we posit that this (natural) multimodal reasoning ability is the most significant reason for human-written explanations being preferred in 48 of the 70 samples in this study (Table~\ref{tab:humor_explanation_choice}). In contrast, the generated explanations often use similar patterns of reasoning: descriptions of absurdity or unexpectedness. We showcase this discrepancy below where the human interpretation (that was preferred) did not explicitly use absurdity or unexpectedness in their reasoning.
% ------------------------------ Example 1 ------------------------------
\begin{figure}[H]
    \centering
    \includegraphics[width=0.82\linewidth]{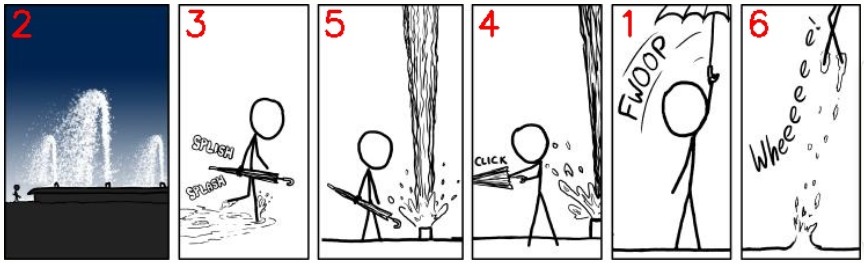}
    \caption{An xkcd comic that requires multimodal reasoning: pairing the sound effects with the comical actions of the stickman to understand how the humor is produced. Access at \url{https://xkcd.com/1038/}}
    \label{fig:xkcd_1038}

    \vspace{0.1em}

    \AlignedExplainRowA
    {The humor in this comic lies in how the stickman "fly away" (in a cute and silly way) with his umbrella. We can observe this by following the narrative where the stickman encounters a powerful water jet at the water fountain. Afterwhich, he opened his umbrella to catch the strong water jet and be propelled upwards. \green{The humor is amplified with the "wheeee" sound effect} by the stickman.}
    {This comic is funny because it \red{subverts expectations} by showing a character using a fountain as a makeshift launch pad with an umbrella, which is an \red{absurd and unexpected use} of everyday objects. The sequence of panels builds up the anticipation, and the final "Wheeee" adds a playful tone to the character's unconventional flight. The humor lies in the combination of creativity and the sheer silliness of the situation }
\end{figure}

% ------------------------------ Example 2 ------------------------------
\begin{figure}[H]
    \centering
    \includegraphics[width=0.3\linewidth]{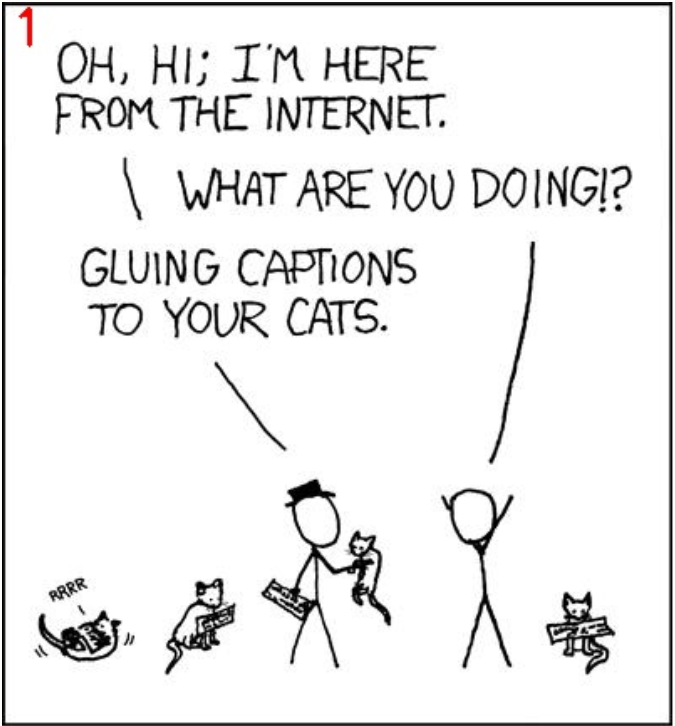}
    \caption{An xkcd comic that requires the user to have relevant world knowledge about the internet, as well as the pairing of expressive phrases such as "What are you doing!?" with the visual expression of the characters (raised hands depicting disbelief) to amplify humor. Access at \url{https://xkcd.com/262/}}
    \label{fig:xkcd_262}

    \vspace{0.1em}

    \AlignedExplainRowB
    {The humor in this comic plays with how the internet likes to put captions on cat related content to give them a more humorous/funny/human expression in the content (memes, short form videos etc). The humor is \green{further amplified by how the other man shouted "what are you doing" with raised hands, showing that he is shocked and/or confused and/or in disbelief} that the man is doing something absurd to the cats.}
    {This comic is funny because it plays on the internet meme culture of adding humorous captions to cat pictures, taking it to an \red{absurd literal level} by physically gluing captions to cats. The \red{unexpected and ridiculous action} contrasts with the normalcy of the conversation, enhancing the humor. Additionally, the cats' reactions add a layer of visual comedy.}
\end{figure}

\noindent\red{Red text} indicate a common reasoning pattern for humor in LMMs.

\noindent\green{Green text} indicate how the human's interpretation uses both modalities to explain the humor.